\documentclass[10pt,twocolumn,letterpaper]{IEEEtran}

\usepackage{times}
\usepackage{epsfig}
\usepackage{graphicx}
\usepackage{amsmath}
\usepackage{amssymb}
\usepackage{subcaption}
\usepackage{booktabs}
\usepackage{gensymb}
\usepackage{nccmath}
\usepackage{array}
\usepackage{multirow}
\usepackage{caption}
\usepackage{placeins}
\usepackage{flushend}
\usepackage{xspace}
\usepackage{adjustbox}
\usepackage{comment}
\usepackage{makecell}

\newcolumntype{F}[1]{%
    >{\raggedright\arraybackslash\hspace{0pt}}p{#1}}%
\newcolumntype{T}[1]{%
    >{\centering\arraybackslash\hspace{0pt}}p{#1}}%
\usepackage{enumitem}
\setlist{nolistsep,leftmargin=*}

\usepackage[pagebackref=true,breaklinks=true,colorlinks,bookmarks=false]{hyperref}

\usepackage[capitalize]{cleveref}
\crefname{section}{Sec.}{Secs.}
\Crefname{section}{Section}{Sections}
\Crefname{table}{Table}{Tables}
\crefname{table}{Tab.}{Tabs.}

\makeatletter
\DeclareRobustCommand\onedot{\futurelet\@let@token\@onedot}
\def\@onedot{\ifx\@let@token.\else.\null\fi\xspace}
\def\etal{\emph{et al}\onedot}

\graphicspath{{img/}} 

\begin{document}
\title{IC3D: Image-Conditioned 3D Diffusion\\for Shape Generation}
\author{%
Cristian Sbrolli \quad Paolo Cudrano\quad Matteo Frosi \quad Matteo Matteucci\\%
Department of Electronics Information and Bioengineering\\%
Politecnico di Milano, Italy \\%
{\tt\small \{name.surname\}@.polimi.it}%
}%
\twocolumn[{%
\renewcommand\twocolumn[1][]{#1}%
\maketitle
\begin{center}
    \centering
    \captionsetup{type=figure}
    \includegraphics[width=\textwidth]{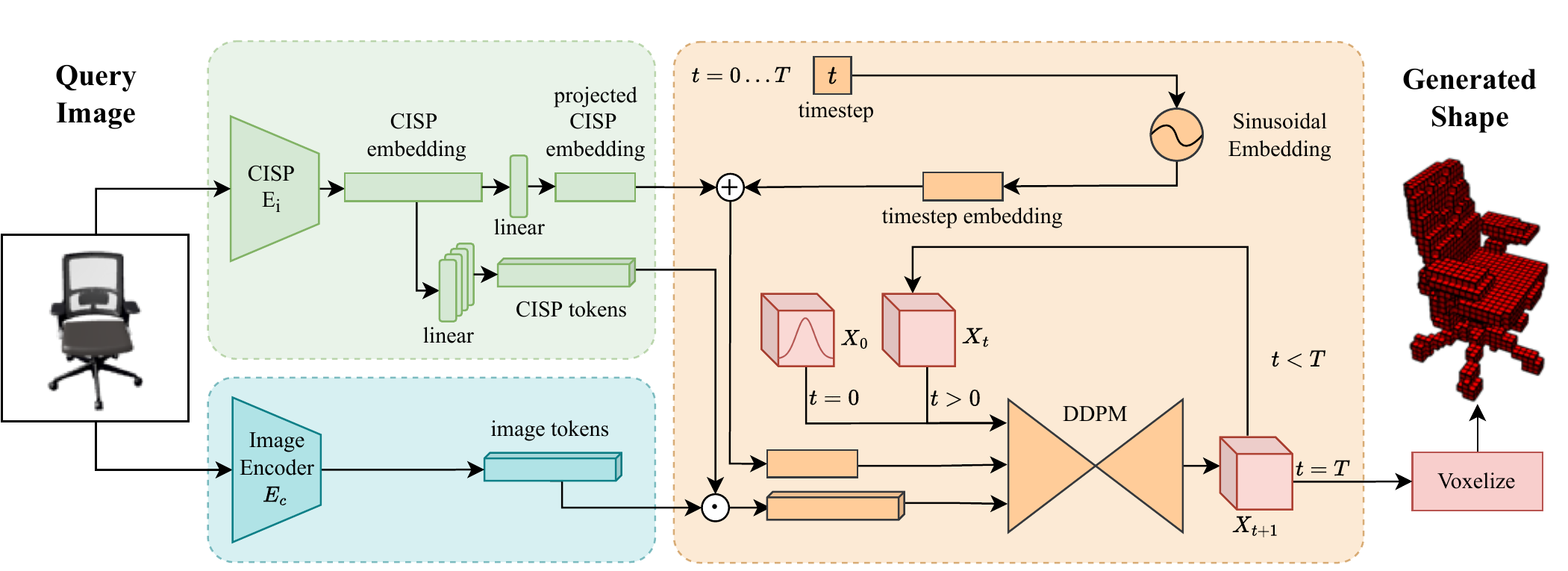}
    \captionof{figure}{IC3D, our 3D shape generation pipeline. A joint image-shape embedding space is preemptively learned through our pre-training model CISP (Contrastive Image-Shape Pre-training). Given a query image, we generate its CISP  embeddings and use them, together with additional context, to condition a 3D Denoising Diffusion Probabilistic Model (DDPM). In this fashion, IC3D can generate diverse 3D shapes, all maintaining both realism and coherence with the query image./\vspace{2em}}
    \label{fig:CISPGuidanceArchitecture}
\end{center}%
}]

\begin{abstract}
   In recent years, Denoising Diffusion Probabilistic Models (DDPMs) have demonstrated exceptional performance in various 2D generative tasks. Following this success, DDPMs have been extended to 3D shape generation, surpassing previous methodologies in this domain. While many of these models are unconditional, some have explored the potential of using guidance from different modalities. In particular, image guidance for 3D generation has been explored through the utilization of CLIP embeddings. However, these embeddings are designed to align images and text, and do not necessarily capture the specific details needed for shape generation. To address this limitation and enhance image-guided 3D DDPMs with augmented 3D understanding, we introduce CISP (Contrastive Image-Shape Pre-training), obtaining a well-structured image-shape joint embedding space. Building upon CISP, we then introduce IC3D, a DDPM that harnesses CISP's guidance for 3D shape generation from single-view images. This generative diffusion model outperforms existing benchmarks in both quality and diversity of generated 3D shapes. Moreover, despite IC3D's generative nature, its generated shapes are preferred by human evaluators over a competitive single-view 3D reconstruction model. These properties contribute to a coherent embedding space, enabling latent interpolation and conditioned generation also from out-of-distribution images. We find IC3D able to generate coherent and diverse completions also when presented with occluded views, rendering it applicable in controlled real-world scenarios.
\end{abstract}
\section{Introduction}
\label{sec:intro}
Recent years have witnessed astonishing advances in deep generative models, especially in image generation. Denoising Diffusion Probabilistic Models (DDPMs) played a central role in this progress, outperforming previous methods such as variational autoencoders (VAEs) and generative adversarial networks (GANs) in unconditional image synthesis~\cite{diffBeatGANs} and text-to-image synthesis. Works such as GLIDE~\cite{GLIDE}, DALLE-2~\cite{DALL-E2}, Stable Diffusion~\cite{stablediffusion}, and Imagen~\cite{Imagen} showed, indeed, how we can effectively condition the generation of images by text prompts.

Transitioning from 2D image generation to 3D shape generation presents challenges, although recent years have witnessed notable progress in this field. Following the initial wave of VAEs and GAN architectures~\cite{3D-VAE, 3D-GAN}, a subsequent wave of models has focused on flow-based and energy-based paradigms~\cite{SoftFlow, PointFlow, 3DDesNet}, showcasing their efficacy in generating high-quality and diverse shapes.
Recent studies have not only demonstrated DDPMs' superiority in shape quality and diversity compared to previous approaches, but have also showcased the potential to condition the generation process using various modalities, such as text prompts~\cite{googleText2Shape}, shape latents from autoencoders~\cite{luo2021diffusion}, and single-view images~\cite{lion}.

In particular, the latter---image-to-shape generation---, is pivotal for driving innovation across diverse fields like VR and AR, cultural heritage, medical imaging and diagnosis, and industrial design and manufacturing. Image-conditioned 3D generation has been experimented~\cite{lion} by leveraging CLIP~\cite{CLIP} image embeddings as guidance tokens. However, the inherent alignment of text and images within CLIP embeddings may not optimally suit guiding an image-to-shape diffusion process, potentially leading to loss of fine-grained details crucial for shape generation.

Motivated by this insight, in this work we propose the construction of a CLIP-inspired joint image-shape embedding space, named CISP (Contrastive Image-Shape Pre-training). We show that this embedding space not only exhibits a well-organized structure but also captures the inherent semantic structural characteristics of the represented objects. 
We exploit this regularity to condition a 3D diffusion model through classifier-free guidance~\cite{cfreeGuidance}, leveraging CISP to steer the generated shapes toward the desired image context.
As our aim is to investigate the advantages of image-shape embeddings for 3D generation, we focus on voxels, to avoid additional architectural complexities due to other representations such as point clouds. 
We find that IC3D generates image-coherent yet diverse shape samples from single-view images, and it achieves cutting edge results for both quality and diversity also in unconditional generation. 
Thanks to the structural properties acquired via CISP guidance, the shapes modeled by IC3D are preferred by human assessors in terms of quality and coherence also when compared to a single-view 3D reconstruction model, despite IC3D being a generative model.
We show the regularity of CISP embeddings through manifold interpolation and OOD probing experiments. Furthermore, we test IC3D's resilience to occluded views and demonstrate its potential for automated deployment in controlled real-world scenarios.

\section{Related Works}
\label{sec:relWorks}

Various strategies have been put forth to develop generative models with the ability to synthesize three-dimensional (3D) objects. Notable methodologies encompass early Variational Autoencoders (VAEs)~\cite{VAE} and Generative Adversarial Networks (GANs)~\cite{GANs} applied to 3D voxelized shapes, as well as the more recent SDF-StyleGAN~\cite{sdfstylegan}, which constructs a generative model based on an SDF-based implicit representation. The research in this domain has also seen a growing interest in generative techniques like flow-based models~\cite{normalizingFlows}, Energy-Based Models (EBMs)~\cite{energybasedtutorial}, and Denoising Diffusion Probabilistic Models (DDPMs)~\cite{ddpmOriginal, ddpm}. Flow-based models, such as DPF-Net~\cite{dpfnet}, manipulate probability distributions to generate samples through variable transformations. EBMs, exemplified by 3D DescriptorNet~\cite{3DDesNet}, optimize energy functions over observed variables and generate new data via Langevin Dynamics~\cite{implicitLangevin, learningMCMC}. On the other hand, DDPMs train using a denoising objective and subsequently invert the process to generate samples from noise.

Point-Voxel Diffusion (PVD)~\cite{PVD} employs point-voxel CNN~\cite{pvCNN} to generate point cloud shapes, also highlighting the inability of training a voxel-based DDPM. Additionally, PVD necessitates separate training for different shape categories due to its unconditional nature. A contrasting approach presented by Luo and Hu~\cite{luo2021diffusion} involves a point-cloud DDPM conditioned on shape latents extracted from a point cloud autoencoder, enabling a single model to generate diverse object categories. Hui~\etal~\cite{NeuralWavelet} employ diffusion on SDFs wavelet coefficients to generate coarse volumes and a refiner network to predict details. Leveraging latent diffusion, LION~\cite{lion} demonstrates image-conditioning of a 3D generation model using CLIP~\cite{CLIP} text or image embeddings. However, although well-structured, CLIP embeddings inherently lack 3D features, and their use in 3D generation might lead to loss of fine-grained structural shape details that are not captured by images and text. On the contrary, joint image-shape embeddings would overcome this limitation.

Joint image-shape embeddings have been investigated by Li~\etal~\cite{li2015joint}, establishing a joint image-shape embedding space through a multistep process. It begins by constructing an embedding space for shapes based on shape similarities and then learns to associate images with their corresponding shape embeddings. In contrast, Kuo~\etal~\cite{kuo2020mask2cad} adopt a holistic approach, jointly learning image and 3D CAD shape embeddings. This concept is further extended in~\cite{kuo2021patch2cad}, learning an image patch-CAD shape mapping. Imagebind~\cite{imagebind} shows instead how we can align multiple modalities to images with a contrastive approach. However, they only focus on depth maps, which do not fully capture the 3D scene and may this lose critical 3D information. On the other hand, we propose to retain full 3D information by extending the contrastive methodology in~\cite{CLIP} directly to image-shape domain, with voxels as 3D representation.

\section{Contrastive Image-Shape Pre-training}
To condition the 3D shape generation on images, we require a meaningful and well-structured concept space jointly encoding both 2D and 3D information. We obtain this space with CISP (Contrastive Image-Shape Pre-training), which learns to embed images and shapes in a joint space. The CISP pre-training method follows the procedure described in CLIP~\cite{CLIP}. In particular, it uses a contrastive approach to learn meaningful representations of images and shapes in a joint space. We define an encoder $E_s$ processing shapes and an encoder $E_i$ processing images, both producing embeddings of size $f$. Given a batch containing $N$ corresponding \textit{(image, shape)} pairs, our training objective is to match each image with its shape. We produce the batches of image and shape embeddings $e_i$ and $e_s$ using $E_i$ and $E_s$. We then compute a symmetric $N \times N$ similarity matrix between image embeddings (rows) and shape embeddings (columns), using a cosine similarity measure. Our training loss is composed of two cross-entropy terms, measuring both the ability to predict the correct shape given an image and vice-versa. This objective aims at maximizing the similarity of the $N$ matching \textit{(image, shape)} pairs and, at the same time, minimizing the similarity of the $N^2-N$ unmatching pairs. For the implementation details, we refer the reader to~\cite{CLIP}.

In CLIP, Radford \etal~\cite{CLIP} adopt a Vision Transformer (ViT)~\cite{Vit} as image encoder. They found it to produce better results and higher computational efficiency compared to CNN architectures when trained on a sufficiently large dataset. As the dataset we use (\cref{sec:experiments}) is limited, however, we employ a Data-efficient image Transformer (DeiT)~\cite{DeiT}, i.e., DeiT Base (DeiT-B), using 768-dimensional hidden embeddings and 12 layers with 12 attention heads each. 

As shape encoder, we extend the ViT model to 3D. We replace the 2D convolutions, mapping an image to patch embeddings, with 3D convolutions mapping a voxel shape to patch embeddings.
A learnable token is prepended to the input sequence, processed by the network, and then projected to the desired embedding dimension. This technique allows us to easily extract the image (or shape) CISP embedding, and is inspired by class tokens first used in NLP~\cite{devlin2018bert} and later introduced in ViT~\cite{Vit}.

\section{Shape Generation Pipeline}
\Cref{fig:CISPGuidanceArchitecture} shows our image-driven shape generation pipeline, which is inspired by 2D text-to-image works~\cite{DALL-E2, GLIDE}. From the input query image we obtain its CISP embedding, carrying information about its 3D shape, and some additional image tokens, encoding information not captured by CISP. We leverage these embeddings to guide our DDPM generative process. As DDPMs are iterative methods, they require information about the current timestep, which is encoded by sinusoidal embeddings~\cite{transformer}. Our DDPM module iteratively refines an input 3D tensor, which is finally transformed into the output shape by binary thresholding. 
We report a detailed description of DDPMs in the supplementary material.

\subsection{Architecture}
\label{sec:DDPMarchitecture}
We extend the architecture used in GLIDE~\cite{GLIDE}, which is based on the ADM model~\cite{diffBeatGANs}. Specifically: (1) we replace 2D convolutions with 3D convolutions; (2) we replace the text encoder with an image encoder $E_c$, which we chose to be a DeiT~\cite{DeiT} model; (3) as in~\cite{DALL-E2}, we use CISP image embeddings in two ways: first, we project and add them to the timestep embedding; second, in each attention block of the network, we project CISP embeddings into four extra tokens and we concatenate them to the attention context (keys, values); (4) we prepend 8 learnable tokens to $E_c$ and use the corresponding outputs as additional attention context, as with CISP embeddings.
\subsection{Guidance}
\label{sec:guidance}
We leverage CISP to train a single DDPM on all the considered object categories, guiding the generation by using CISP image embeddings. To this end, we train our model with classifier-free guidance~\cite{cfreeGuidance}, which eliminates the need for a separate classifier and obtains similar results compared to classifier guidance~\cite{ddpm}. We report the mathematical formulation of classifier-free guidance and guidance scale analysis in the supplementary material.

\section{Experiments}
\label{sec:experiments}
We run experiments to evaluate our generative pipeline quantitatively and qualitatively. We compare our model against state-of-the-art 3D generative approaches to quantitatively evaluate our generated shapes, following evaluation approaches and metrics defined by the literature. Furthermore, we evaluate the effectiveness of our guidance approach and the visual quality of generated shapes through a human evaluation experiment. Lastly, we investigate the representation and generalization capabilities of CISP and IC3D through additional experiments and ablations.

Following previous shape generation works~\cite{PVD,luo2021diffusion,PointFlow,SoftFlow,lion}, we focus on the Airplane, Car, and Chair categories from the ShapeNet~\cite{shapeNet} test set. Although not used in previous works, we also train our model on the Table and Watercraft categories; see the supplementary material for the results of these two additional categories. Notice that, except for the unconditional experiments, we refer to IC3D as a single model trained on all categories.

\begin{figure}[t]
  \centering
   \includegraphics[width=\linewidth]{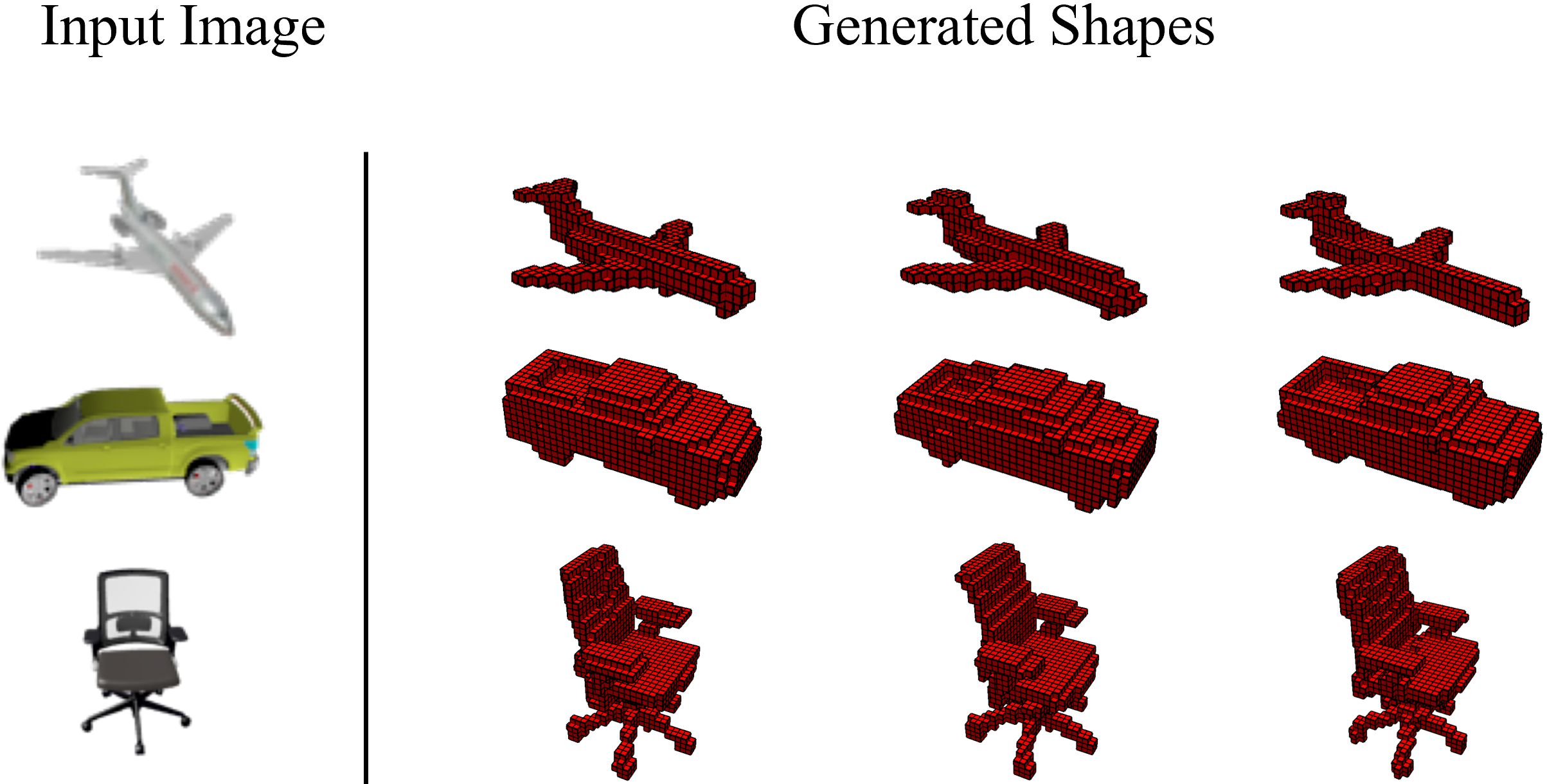}
   \caption{3D shapes generated by IC3D from the query images on the left. Notice how each shape is diverse and yet it coherently represents structural elements of the query (e.g., flat-topped tail of the airplane, cargo space of the pickup truck, wheels and armrests of the chair.)\vspace{-.5em}}
   
   \label{fig:generationExamples}
\end{figure}

\subsection{Shape Generation}
The generation of shapes is performed using the inverse diffusion process. In particular, we first generate a pure noise sample of dimension $32^3$, and then run 1000 backward diffusion steps.
We obtain a $32^3$ float tensor which we voxelize through binary thresholding (threshold at 0.5). For conditional generation, we apply classifier-free guidance as explained in \cref{sec:guidance}, with a guidance factor of $1.5$, as we found it produces the best results. \Cref{fig:generationExamples} shows examples of image-guided generation from our model.

\begin{table}[t]
  \centering
  \begin{tabular}{@{}l l c c@{}}
    \toprule
    ~ & ~ & \multicolumn{2}{c}{1-NNA(\%)}\\
    Shape & Model & CD & EMD \\ 
    \midrule
    ~ & Shape-GF~\cite{ShapeGF} & 80.00 & 76.17 \\
    ~ & DPF-Net~\cite{dpfnet} & 75.18 & 65.55 \\
    ~ & \cite{NeuralWavelet} & 71.69 & 66.74 \\
    Airplane & \cite{luo2021diffusion} & 62.71 & 67.14 \\
    ~ & PVD~\cite{PVD} & 73.82 & 64.81 \\
    ~ & LION~\cite{lion} & 67.41 &  61.23 \\
    ~ & \textbf{Ours} & \textbf{58.93} & \textbf{56.93} \\ 
    \midrule
    ~ & Shape-GF~\cite{ShapeGF} & 63.20 & 56.53 \\
    ~ & DPF-Net~\cite{dpfnet} & 62.35 & 54.48 \\
    ~ & \cite{NeuralWavelet} & - & - \\
    Car & \cite{luo2021diffusion} & - & - \\
    ~ & PVD~\cite{PVD} & 54.55 & 53.83 \\
    ~ & LION~\cite{lion} & 53.70 & \textbf{52.34} \\
    ~ & \textbf{Ours} & \textbf{53.20} & 53.11 \\ 
    \midrule
    ~ & Shape-GF~\cite{ShapeGF} & 68.96 & 65.48 \\
    ~ & DPF-Net~\cite{dpfnet} & 62.00 & 58.53 \\
    ~ & \cite{NeuralWavelet} & 61.47 & 61.62 \\
    Chair & \cite{luo2021diffusion} & 62.08 & 64.45 \\
    ~ & PVD~\cite{PVD} & 56.26 & 53.32 \\
    ~ & LION~\cite{lion} & 53.41 & \textbf{51.14}\\
    ~ & \textbf{Ours} & \textbf{53.30} & 51.97 \\ 
    \bottomrule
\end{tabular}
\caption{Comparison of unconditional generation performance of IC3D against baselines and SoTA 3D generative models. Best results are highlighted in bold. 1-NNA perfect score is 50\%. The car category was not evaluated in~\cite{luo2021diffusion} and~\cite{NeuralWavelet}.}
\label{tab:quantitativeResults}
\end{table}

We evaluate the quality and diversity of shapes generated by our model by comparing against other 3D generative works~\cite{ShapeGF,dpfnet,NeuralWavelet,luo2021diffusion,PVD,lion} and we report results in \cref{tab:quantitativeResults}. Since these models only evaluate unconditional generation, we also configure our pipeline to match this setting for a fair comparison. To achieve unconditional generation with our model, we conduct experiments on each category without guiding images, substituting CISP embeddings and $E_c$ tokens with learned null tokens (only at test time). We compare mainly on 1-Nearest Neighbor Accuracy (1\nobreakdash-NNA), as it was shown to be the most representative metric for 3D generation~\cite{PointFlow}. We report in the supplementary material results for other common generation metrics, i.e., 
Minimum Matching Distance (MMD), Coverage (COV) and Shading-image-based FID~\cite{sdfstylegan}, where IC3D also obtains competitive results. For each metric, we consider both Chamfer distance (CD) and Earth Mover's Distance (EMD).

1\nobreakdash-NNA, presented in~\cite{1nna} and adopted for 3D generation in~\cite{PointFlow}, measures both generated samples diversity and quality by considering the accuracy of a 1-Nearest Neighbor classifier. We define $S_g$ as the set of generated samples and $S_r$ as the set of reference samples (which we take from the test set) with $|S_r|=|S_g|$. 
Also, let $N_x$ be the nearest neighbor of a sample $x$, with ${N_x \in \{S_g \cup S_r - x\}}$. 1\nobreakdash-NNA is then defined as

\begin{equation}
    \medmath{\operatorname{1-NNA}(S_g, S_r) = \frac{\sum\limits_{x \in S_g} \mathbb{I}[N_x \in S_g]+\sum\limits_{x \in S_r} \mathbb{I}[N_x \in S_r]}{|S_g|+|S_r|}},
\end{equation}
with $\mathbb{I}$ being the indicator function. As we are measuring the accuracy of a 1\nobreakdash-NN classifier on recognizing generated and reference shape, a value of 50\% is the perfect score.

\begin{figure}[t]
  \centering
   \includegraphics[width=0.8\linewidth]{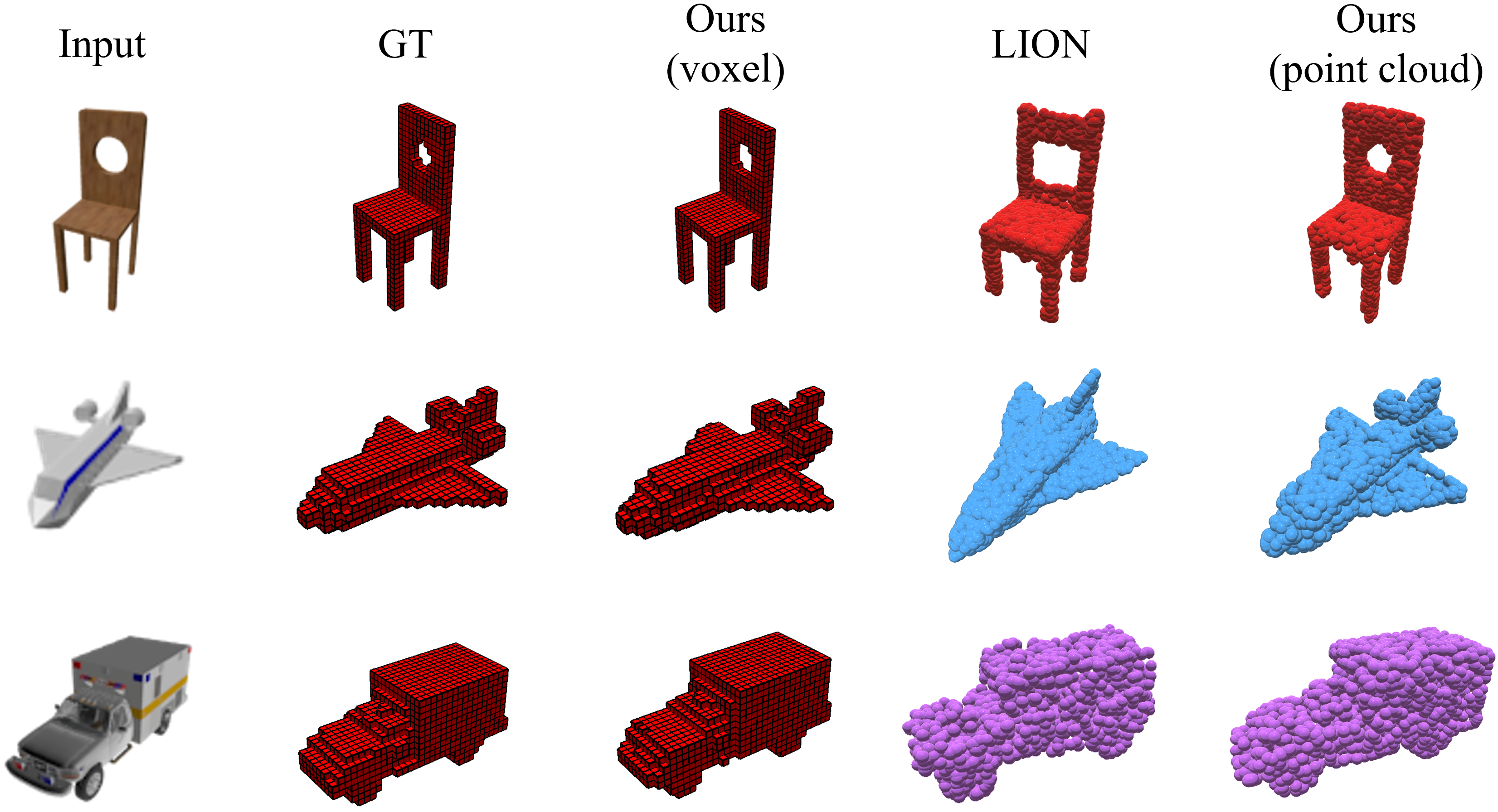}
   \caption{Thanks to CISP guidance, our model is able to match finer-grained details of the guiding image with respect to LION~\cite{lion}, which uses CLIP. For our model, we show generated voxels and the corresponding sampled point cloud.}
   \label{fig:visualComparisonPCVoxel}
\end{figure}

To make voxel shapes compatible with EMD and CD, we follow the approach from PVD~\cite{PVD}, sampling 2048 points from the surface of the generated shape. We also adopt metrics implementations from PVD~\cite{PVD} public code. Our method outperforms or is on par with other models in terms of quality and diversity, achieving higher results on the airplane class and comparable results with LION~\cite{lion} on cars and chairs. As LION~\cite{lion} also allows for image guidance, but only evaluates on unconditional generation, we furhter compare these two models qualitatively in \cref{fig:visualComparisonPCVoxel}. Thanks to CISP guidance, our model is better able to generate shapes that are coherent with the input image in most details, as we can observe with the plane back engines and the chair circular hole.

We further analyze the quality of our results running a \textbf{side-by-side human evaluation}, comparing our model with a SoTA 3D reconstruction approach. We evaluate how well our generated shapes match query images, as well as their realism. Each human evaluator is presented with two unlabeled and randomly ordered 3D shapes, one from our model and one from a reconstruction model. The evaluator is asked to choose which of the two shapes is more realistic. After its choice, the query image is shown, and the evaluator has to choose which shape is more coherent with the displayed image. We chose to only show the query image after the first question is answered to avoid biasing the evaluator when selecting the most realistic shape. Shapes are displayed as GIFs rotating $360\degree$ around their volume, allowing the evaluator to see each shape from different perspectives. Following the method from~\cite{humanEvaluationMethod}, we show each pair of images exactly to 5 independent annotators, allowing us to reduce the variance on each pair by majority voting. 

We compare against 3D\nobreakdash-RETR~\cite{3d-retr}, the best-performing publicly available model for single-view 3D reconstruction at the time of the experiment. We select 200 random images per category (airplane, car, chair) and use them to condition our model and to predict the shapes using 3D\nobreakdash-RETR, obtaining the pair of shapes needed for the evaluation. We publicly release the images, GIFs, and necessary setup for reproducing this evaluation experiment.

\begin{figure}[t!]
  \centering
   \includegraphics[width=0.8\linewidth]{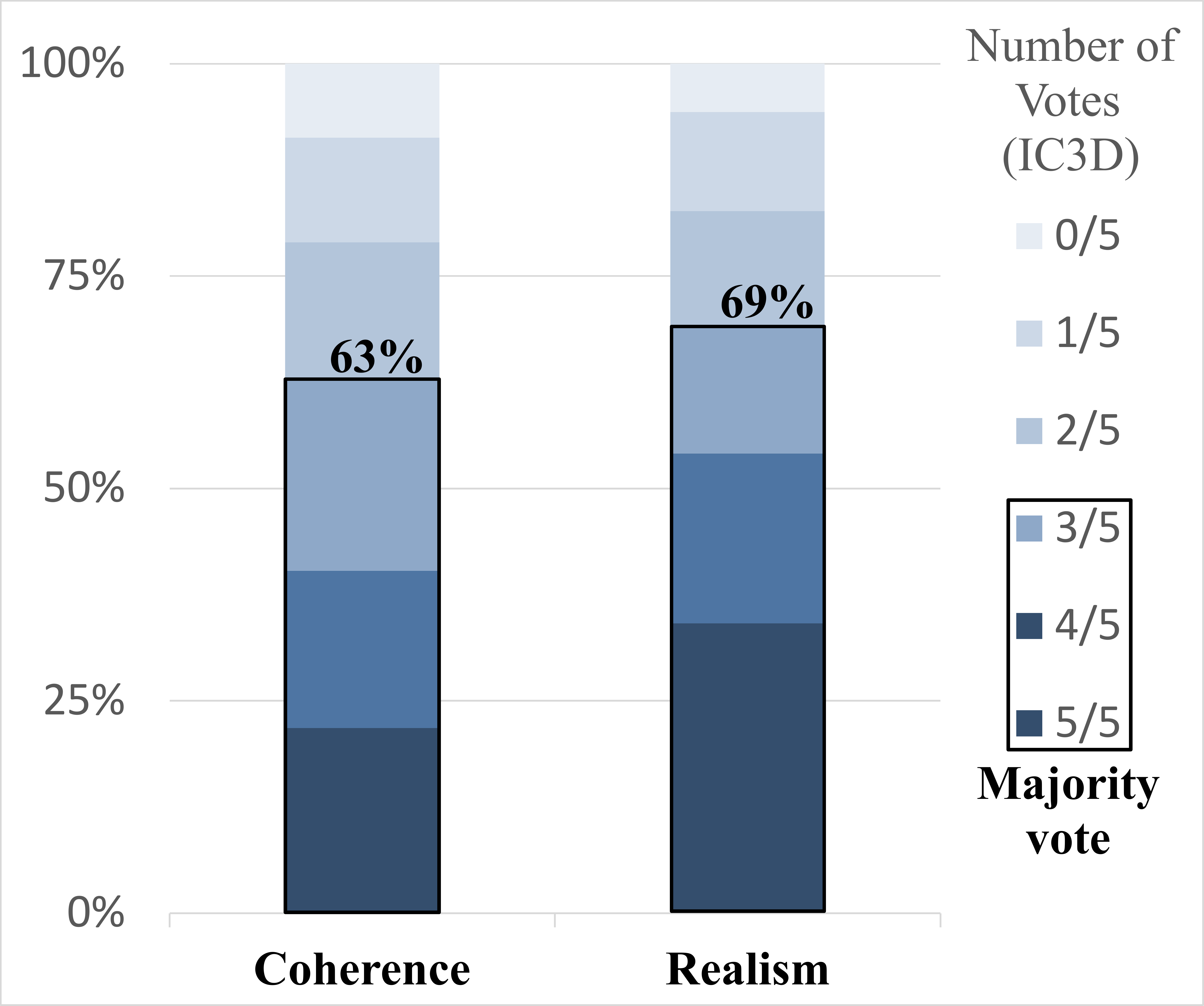}
   \caption{Human evaluation of coherence and realism of IC3D generated shapes against shapes reconstructed by 3D\nobreakdash-RETR~\cite{3d-retr}.
   IC3D shapes were considered more coherent with the input image $63\%$ of the times and more realistic $69\%$ of the times.
   Among the unanimous decisions (5/5 or 0/5 votes), $72\%$ were in favor of IC3D for its coherence and $85\%$ for its realism.}
   \label{fig:heResults}
\end{figure}

The results obtained (\cref{fig:heResults}) show that our model is preferred by human evaluators in both coherence and realism. In particular, the majority of the evaluators considered our generated shapes to be the most coherent with the query image in 62\% of the cases, and  the most realistic in 69\% of the cases. Examining the per-category scores (reported fully in the supplementary material), chairs are the most preferred category for our model in both coherence (68\%) and realism (91\%). 
We speculate that our model achieves better performance because its generative nature forces it to capture the semantics of each structural detail, while for a reconstruction method any voxel has an equivalent meaning, as further discussed in \cref{sec:reconstrabilities}. 

\subsection{Occluded and partial image views}
\begin{figure}[t]
  \centering
   \includegraphics[width=0.8\linewidth]{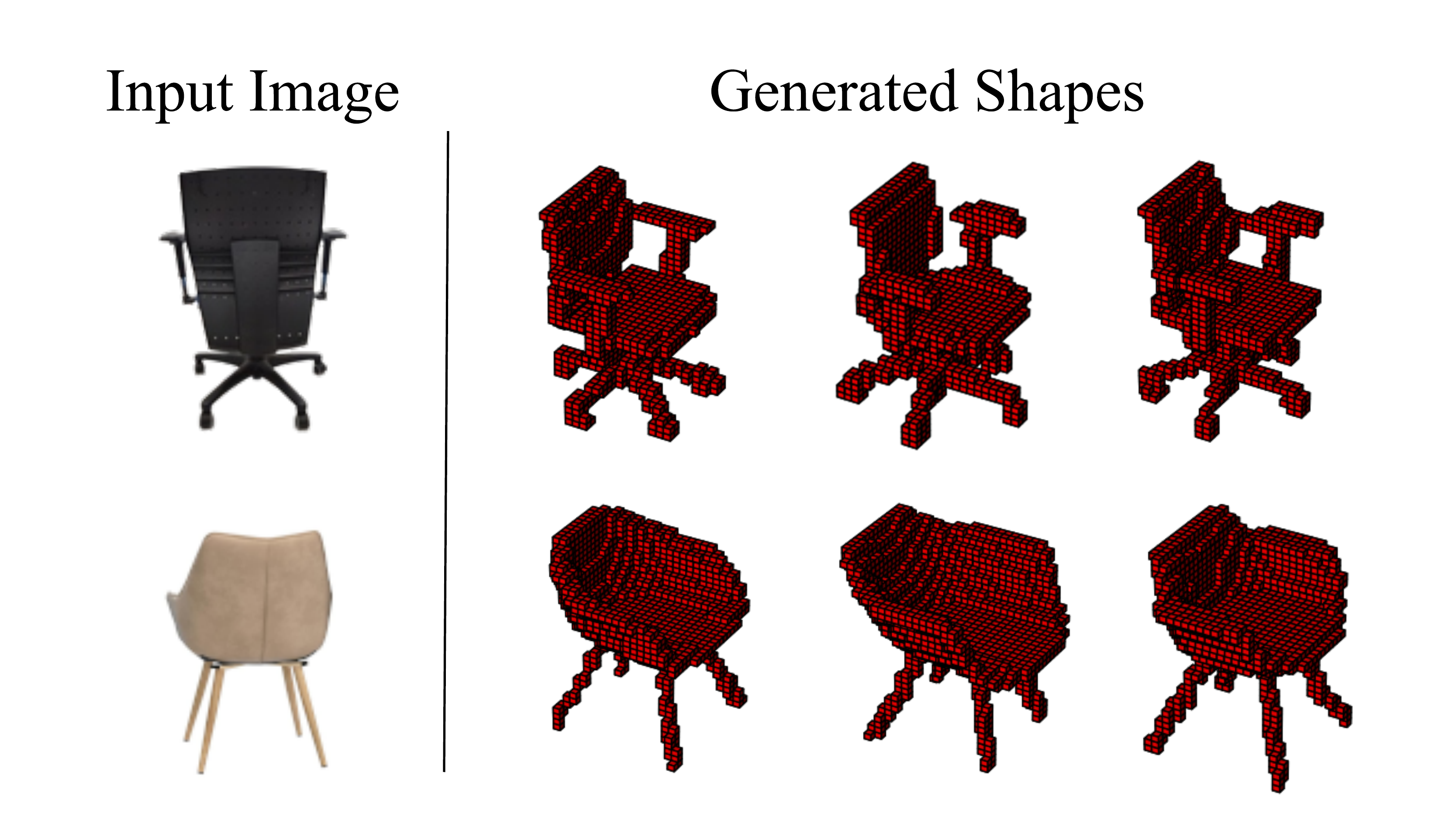}
   \caption{Generation from occluded views. Our model correctly generates plausible and realistic shapes even for unseen parts.}
   \label{fig:ambViews}
\end{figure}

We study the generative power of IC3D when the conditioning image presents occlusions or partial views. In these situations several completions for the occluded parts are possible. The generative nature of IC3D allows it to produce various shapes, each coherent with the given query image, but presenting diversity in the unseen parts. This is evident in \cref{fig:ambViews}, where only backs of chairs are shown to our model. Despite the missing information, IC3D is capable of (1) identifying the type of object it is presented with (as chairs), and (2) generating chairs with plausible but diverse front parts.

\begin{figure}[t]
  \centering
   \includegraphics[width=1\linewidth]{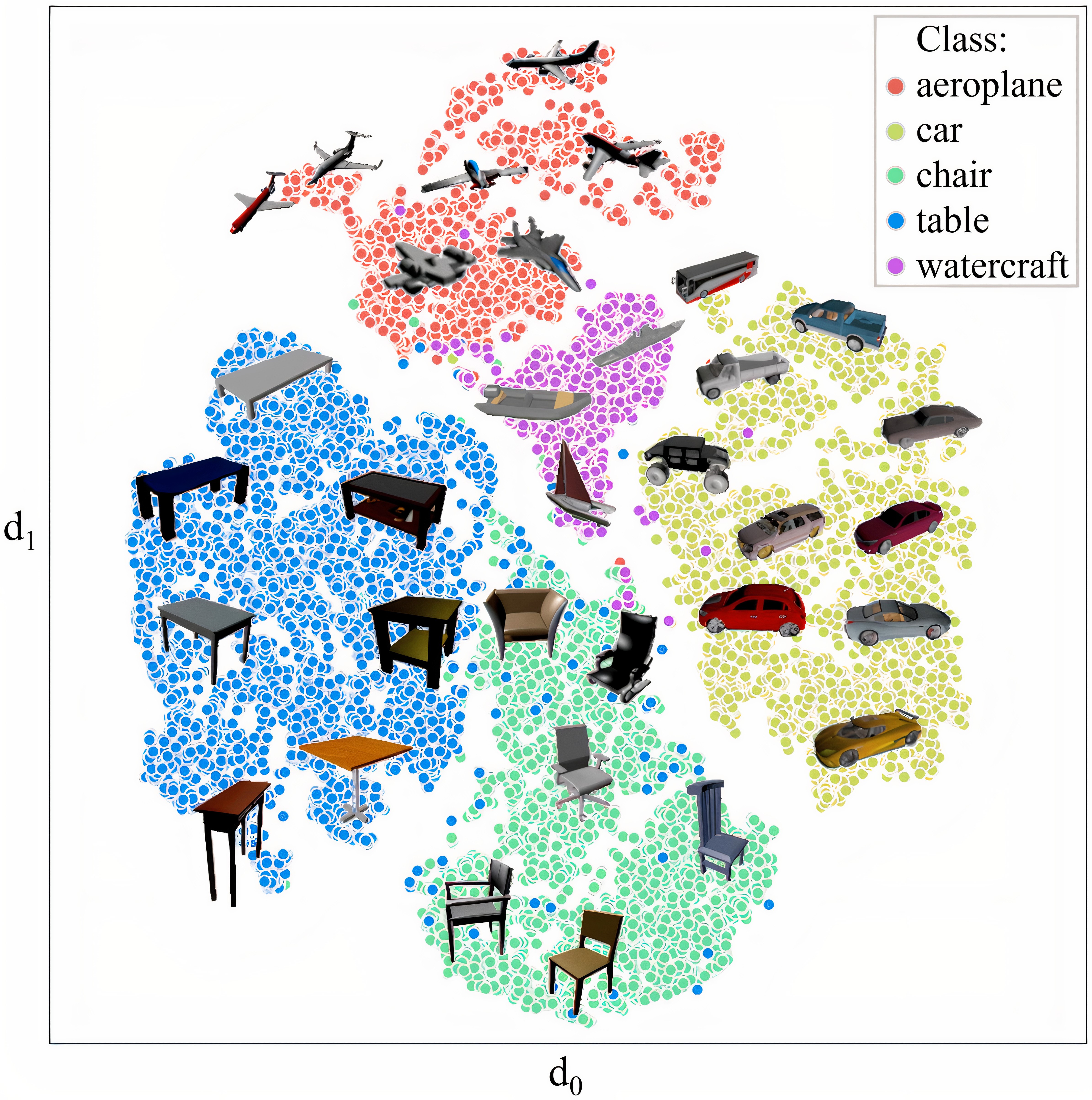}
   \caption{Projection of the CISP shape embeddings into a 2D space, displaying instances of the encoded images for each considered class at their respective locations.\vspace{-.8em}}
   \label{fig:learntSpaceCISP}
\end{figure}

\subsection{Latent space analysis}
To gain insight into how CISP embeddings encode meaningful and effective representations, we analyze the learned latent space. We project CISP embeddings by first applying PCA, reducing them to 50 dimensions, and then using t\nobreakdash-SNE to project into the 2D plane. \Cref{fig:learntSpaceCISP} shows the projected shape embeddings of the training set, in which we observe that the categories are well separated. The figure also shows example images depicting how shapes are organized in the embedding space. We observe how the model captures details and subcategories, allowing it to better structure the space. For example, notice how the height of the tables increases as $d_1$ decreases or how shelves begin to appear as $d_0$ increases. At the same time, for cars, notice how sports cars fade into city cars and then into progressively larger cars as $d_1$ increases. We also see inter-class interactions, e.g., the small cloud of buses lies close to longest watercrafts since they share a similar form factor when voxelized. This well-structured joint space makes CISP the core of our guidance method, allowing us to generate shapes coherent with input images.

\subsection{Sketch-to-Shape generation}
\label{sec:handDrawn}
\begin{figure}[t]
  \centering
  \includegraphics[width=0.8\linewidth]{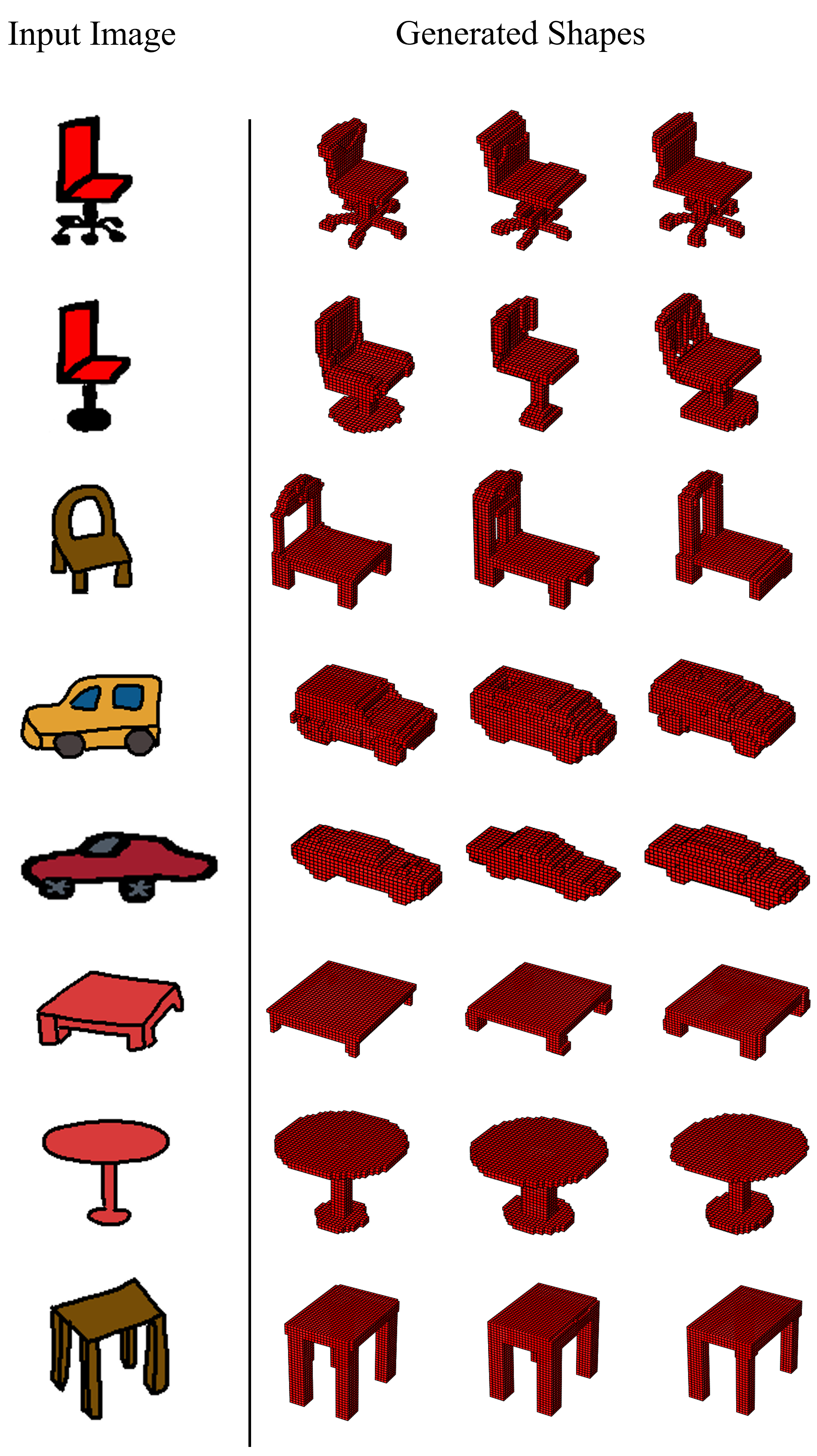}
   \caption{The generalization power of CISP allows us to generate coherent shapes even when feeding IC3D with a hand-drawn sketch of the desired object. Notice how the generated samples accurately capture characteristic aspects of the drawn items, such as the type of chair bottom (wheels, single leg, four legs), while still providing a degree of diversity.\vspace{-1em}}
   \label{fig:handDrawn}
\end{figure}
To additionally prove the effectiveness and generalization capabilities of the CISP model, we test the generation of 3D shapes by guiding our DDPM with CISP embeddings of hand-drawn sketches. Examples of sketch-to-shape generation are shown in \cref{fig:handDrawn}. CISP embeddings correctly encapsulate shape categories and even more specific intraclass details from the hand-drawn sketches.

\subsection{Interpolations}
\label{sec:interpolations}
\begin{figure*}[t]
  \centering
   \includegraphics[width=0.8\linewidth]{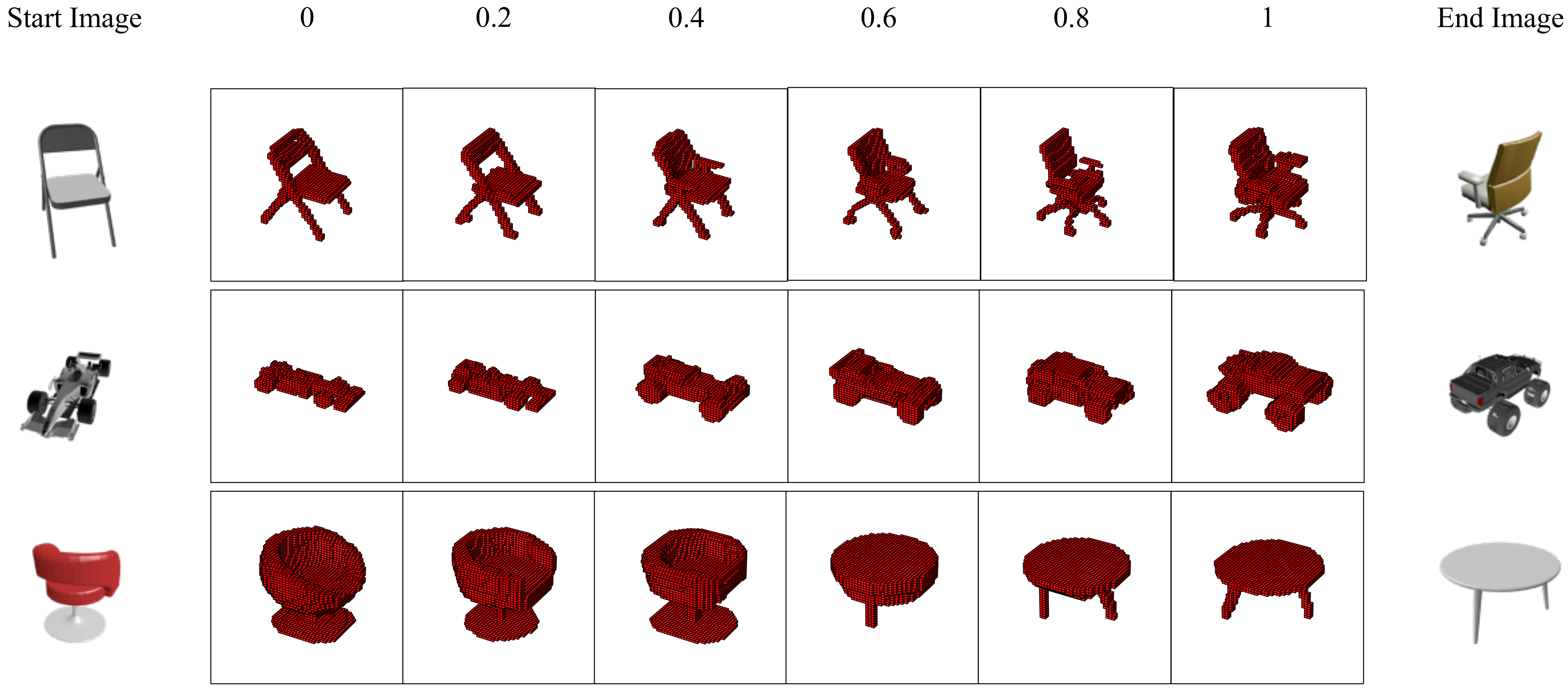}
   \caption{The regularity of our CISP concept space allows IC3D to generate consistent shapes even when interpolating between two unrelated embeddings. Notice how the intermediate shapes mutate smoothly while maintaining perfect structural realism throughout the transformation. At each interpolation step (from $0.0$ to $1.0$), only a few, crisp structural details are changed, behavior that we observe indifferently in intra-class (rows 1--2) and inter-class (row 3) interpolations.}
    \label{fig:interpolations}
\end{figure*}
We show that it is possible to traverse the concept space between two objects by interpolating between their CISP embeddings. As in~\cite{DALL-E2}, we do so by moving between the two considered CISP embeddings with Slerp (spherical linear interpolation). We found Slerp to work much better than standard linear interpolation, which failed to effectively blend the two input shapes. \Cref{fig:interpolations} shows some examples of interpolated shapes obtained with our model. We show that both intra- and inter-class interpolations are effective, further proving the relevance of the knowledge captured by the CISP embedding space. Notice in the first row how the wheels and armrests are added to the original chair structure before turning it into an office chair. Also notice how, in an inter-class scenario (bottom row), the chair is first filled, then the base is removed, and finally the legs are added, completing its transformation into a table.
\subsection{Ablation study}
\begin{table}[!t]
  \begin{adjustbox}{width=\linewidth,center}
  \begin{tabular}{@{}cc|cccc@{}}
    \toprule
    \multicolumn{2}{c|}{Model guidance} & \multicolumn{2}{c}{1\nobreakdash-NNA(\%)} & \multicolumn{2}{c}{COV(\%)$\uparrow$} \\ 
    CISP & $E_c$ & CD & EMD & CD & EMD \\
    \midrule
    Yes & No & 68.71 & 62.37 & 51.49 & 52.35 \\
    Null token & Null token & 55.14 & 54.00 & \textbf{51.99} & \textbf{52.70} \\
    Yes & Null token & 55.21 & 52.90 & 51.14 & 52.59 \\
    Yes & Yes & \textbf{54.55} & \textbf{52.43} & 50.72 & 52.33 \\
    \bottomrule
    \end{tabular}
    \end{adjustbox}
    \caption{We analyze the impact of different guidance elements in IC3D, testing whether removing completely a guidance encoder or substituting it with the null token affects the generative results. Scores are averaged over the three considered categories. We notice that while coverage remains similar in all cases, 1\nobreakdash-NNA score degrades when $E_c$ is removed.\vspace{-.4em}}
    \label{table:ablationresults}
\end{table}

To better understand the contribution of different elements in our model, we perform an ablation study on each of them and report the main results in \cref{table:ablationresults}. We evaluate the generation performance of IC3D when $E_c$ is completely removed and when guidance tokens from CISP and/or $E_c$ are substituted, at inference time, with a null token. For configurations using guidance images (when not using null tokens), we use images from a different set than $S_r$. Unlike text-to-image models~\cite{DALL-E2}, where the extra text encoder has a negligible role, our model suffers a significant drop in performance without $E_c$. Specifically, in this case, the generated shapes retain their diversity, but their visual quality deteriorates considerably. On the other hand, using null tokens has a minor effect on generation quality, while achieving the highest coverage scores. This is expected, as the model is freely generating shapes. These results indicate that CISP and $E_c$ not only offer effective guidance, but also help to learn to generate higher quality samples during training. Moreover, we notice that the guidance from CISP and $E_c$ enables a tradeoff between quality (1-NNA) and diversity (COV).

\begin{figure}[t]
  \centering
   \includegraphics[width=1\linewidth]{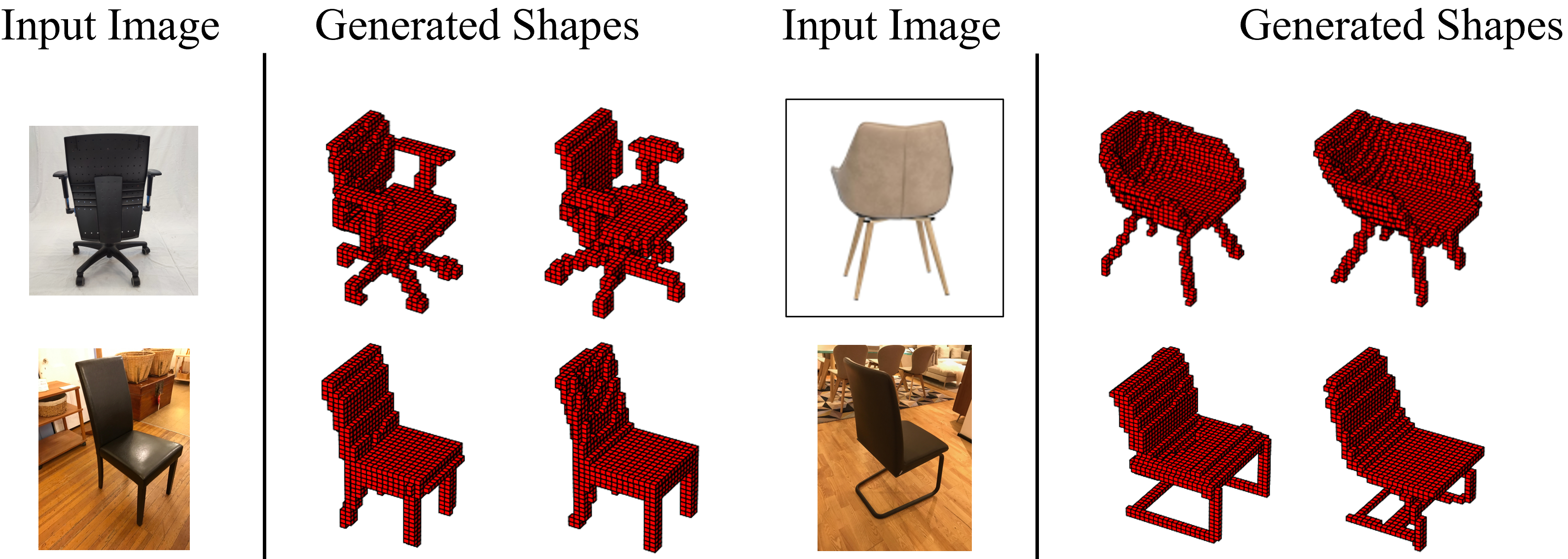}
   \caption{Generation from smartphone photo in a controlled environment and catalogue photo (first row) and from pix3d images (second row). Backgrounds are removed by an automatic tool before feeding them to IC3D, which is able to generate coherent and realistic shapes in all these settings.}
   \label{fig:iiw}
\end{figure}

\begin{table*}[t]
  \begin{adjustbox}{width=\linewidth,center}
  \begin{tabular}{@{}lccccccccccc@{}}
    \toprule
    ~ & \thead{3D\nobreakdash-R2N2\\\cite{3dr2n2}} & \thead{OGN\\\cite{ogn}} & \thead{Pixel2Mesh\\\cite{pixel2mesh}} & \thead{IM\nobreakdash-Net\\\cite{imnet}} & \thead{{Pix2Vox++/F}\\\cite{pix2vox++}} & \thead{3D\nobreakdash-RETR\\\cite{3d-retr}} & \thead{TMVNet \\\cite{tmvnet}} & \thead{Ours\\(1)} & \thead{Ours\\(5)} & \thead{Ours\\(10)} & \thead{Ours\\(15)} \\
    \midrule
    IoU & 0,592 & 0,633 & 0,554 & 0,701 & 0,664 & 0,719 & 0,761 & 0,579 & 0,633 & 0,649 & 0,658 \\ 
    F-Score & 0,377 & 0,409 & 0,416 & 0,445 & 0,485 & 0,492 & 0,572 & 0,363 & 0,402 & 0,414 & 0,421 \\
    \bottomrule
  \end{tabular}
  \end{adjustbox}
  \caption{Reconstruction performance of IC3D against 3D reconstruction models. To evaluate our reconstruction performance, we sample $n$ shapes ($n$ in brackets) and report the maximum reconstruction metric obtained. Being IC3D a generative model, its reconstruction score increases as we sample more shapes.}
    \label{tab:rec_results}
\end{table*}

\subsection{Generation from in-the-wild images}
Although our model is trained only on synthetic data (ShapeNet), we study its ability to generate shapes from in-the-wild images. To perform such analysis in an exhaustive way, we provide IC3D with images coming from three sources, each more in-the-wild: (1) an online chair catalog, (2) a known real-world image-shape dataset (Pix3D~\cite{pix3d}), and (3) pictures of an office chair captured by the authors with a smartphone.
As the samples in our training dataset, ShapeNet, are rendered only on white backgrounds, the input images to our model are pre-processed with a simple and completely automatic background removal tool. 
\Cref{fig:iiw} highlights how our model is capable of generating coherent 3D shapes even in these situations, demonstrating potential for automatic use in controlled environments, such as industrial settings.
As already seen in \cref{fig:ambViews},
also in this scenario IC3D introduces a level of coherent variability to the generated shapes when occlusion and partial image views are given, making it even more appealing for applications in graphics and design.

\subsection{3D Reconstruction abilities}
\label{sec:reconstrabilities}
\begin{figure}[t]
  \centering
   \includegraphics[width=\linewidth]{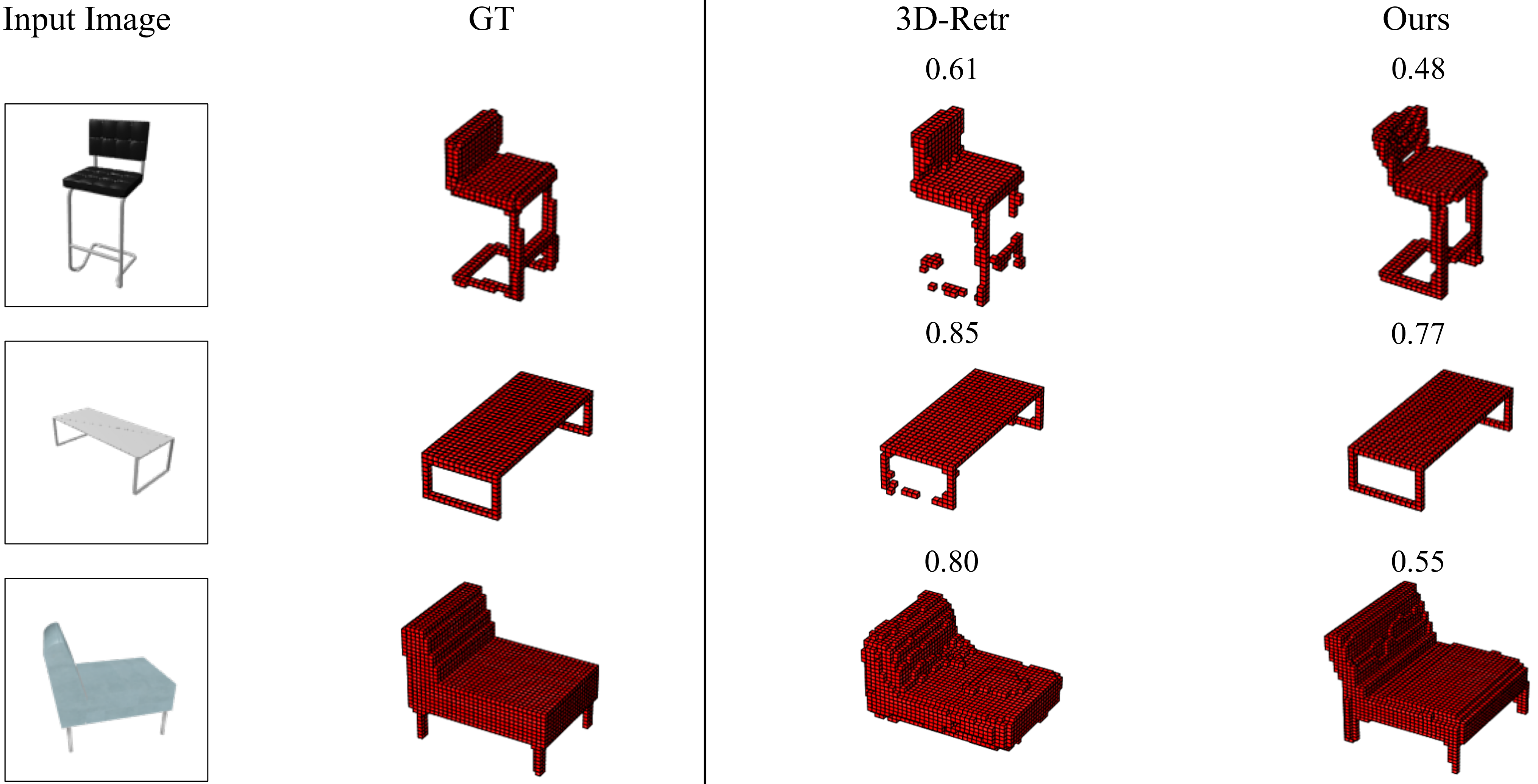}
   \caption{We investigate the application of IC3D to 3D reconstruction, comparing its generated shapes to the reconstructions of 3D\nobreakdash-RETR~\cite{3d-retr}. Being optimized on mere reconstruction losses, 3D\nobreakdash-RETR achieves higher reconstruction scores~(IoU); however, it fails to capture small details, such as the thin legs of a tall chair, which are instead well modeled by IC3D.
   }
   \label{fig:chairExample}
\end{figure}

We investigate the ability of our model to perform single-view 3D shape reconstruction. We compare our generated shapes against several baselines~\cite{3dr2n2,ogn,pixel2mesh,imnet,pix2vox++} and SoTA works~\cite{3d-retr,tmvnet}, evaluating their Intersection over Union (IoU) and F-score. Since IC3D is a generative model, we display the maximum scores obtained when sampling an increasing amount of shapes.
Results, presented in \cref{tab:rec_results}, show that our generative model achieves a satisfactory performance, on par with all baselines, even when used for 3D reconstruction, although this was not our target. \Cref{fig:chairExample} shows how our shapes are more structurally correct and thus realistic, although this is often not reflected in the reconstruction metrics (which focus only on voxel correspondence).

\section{Discussion and limitations}
The experiments presented in the previous sections highlight the ability of our pipeline to produce structurally realistic and diverse shapes, correctly capturing the input data distribution. They also prove the applicability of DDPMs for the image-driven generation of voxel-based shapes. To this end, we presented CISP, a model exploiting contrastive training to learn a joint embedding space for images and shapes. We leveraged its well-structured embedding space to guide IC3D generation toward query images. 
Our quantitative results showed the effectiveness of IC3D in generating high-quality shapes. Our human evaluation survey also confirmed that IC3D achieves higher structural realism than a competitive 3D reconstruction model while maintaining consistency with the query image. We speculate that the generative nature of our model forces it to learn the structural semantics of the training data distribution, to be able to generate diverse but coherent samples. Instead, most reconstruction methods focus only on minimizing a reconstruction metric, without giving any importance to the realism, integrity, and structural correctness of the produced shape. For this reason, given for example the task of reconstructing a chair with thin legs (\cref{fig:chairExample}), if a reconstruction model were to omit an entire leg, its reconstruction score would not be significantly impacted, but the chair would certainly fall over. The generative nature of our model and the usage of a well-structured joint image-shape embedding space, instead, enforces the inductive biases necessary to learn the importance of producing stable chair legs.

The main limitation of IC3D resides in its low sampling speed, an issue that is well-known for simple DDPM models as the one we employ. Several techniques were recently presented to obtain a substantial speed-up~\cite{nichol2021improved, watson2021learning, ddim, latentDiff}, and further research could be invested into integrating them within our pipeline. 
In order to explore the capabilities of CISP-guided DDPMs for image-conditioned shape generation, we developed IC3D focusing on the simplest 3D representation: voxels. This allowed us to decouple the complexities due to the representation (e.g., point clouds~\cite{pointNet, pvCNN}) from the capabilities of the model. We are aware, however, of the limitations of voxels, in particular regarding their high scaling cost. Future extensions will expand our work to produce more complex 3D data, such as point cloud or implicit functions.

Experiments in \cref{sec:handDrawn,sec:interpolations} highlight the importance of the CISP embedding space for producing high quality realistic results.
In its current form, CISP was trained on ShapeNet, a dataset well-known but limited in size. Training CISP on larger datasets would allow for a stronger generalization power, also improving its effectiveness as a zero-shot model. Furthermore, the extension of our single-view pipeline to multi-view guidance through CISP embeddings represents an exciting research direction for future work.

{\small
\bibliographystyle{IEEEtran}
\bibliography{bib}
}

\end{document}


\setcounter{page}{10}
\title{Supplementary Material}
\maketitle
\suppressfloats
\renewcommand{\thesection}{\Alph{section}}

\section{Diffusion Models}
\label{sec:diffusionFormulation}
Denoising Diffusion Probabilistic models~\cite{ddpm, ddpmOriginal} are latent variable models consisting of two processes: the forward noising process \textit{q} and the backward denoising process \textit{p}. We define $x_0$ as the (clean) input data and $x_1, ...,x_T$ as the latent variables sampled at each noising step, each with the same dimensionality as $x_0$. The forward process is a fixed Markov chain gradually adding noise to the data with a variance schedule $\beta_1,...,\beta_T$. Each forward step is then a Gaussian transition following ${q(x_t|x_{t-1}):=\mathcal{N}(x_t;\sqrt{1-\beta_t}x_{t-1};\beta_t\mathbf{I})}$. With a well-designed variance schedule and a sufficiently large $T$, we get that $x_T$ converges to an isotropic Gaussian distribution. Also the backward process is parametrized by a Markov chain with Gaussian transitions, but it is learned instead, and the initial state is sampled from ${x_T \sim \mathcal{N}(x_T; 0; \mathbf{I})}$. Backward transitions follow ${p(x_{t-1}|x_t):=\mathcal{N}(x_{t-1};\mu_\theta(x_t, t);\Sigma_\theta(x_t,t))}$, where $\theta$ are the model parameters. A crucial property of this formulation is that the defined forward process allows us to directly sample from any given timestep. In fact, defining $\alpha_t=1-\beta_t$ and $\Bar{\alpha}_t=\prod_{s=0}^t\alpha_s$ we can write the marginal at timestep $t$ as
\begin{equation}
\label{eq:forwardSample}
    q(x_t|x_0) = \mathcal{N}(x_t;\sqrt{\Bar{\alpha}_t}x_0;(1-\Bar{\alpha}_t)\mathbf{I}),
\end{equation}
which allows us to sample $x_t$ as
\begin{equation}
    x_t = \sqrt{\Bar{\alpha}_t}x_0 + \sqrt{1-\Bar{\alpha}_t}\epsilon, \;\;\; \epsilon\sim\mathcal{N}(0,1).
\end{equation}

\section{Training Approach and Sample Generation}
\label{sec:generationApproach}
As $q$ and $p$ form a Variational Auto-Encoder~\cite{VAE}, we can write a variational lower bound (VLB)~\cite{VAE} and use it as objective for the model. However, Ho \etal~\cite{ddpm} propose a reparametrization of $\mu_\theta(x_t,t)$ allowing a simpler objective and better sample quality with respect to optimizing the VLB. Instead of predicting $\mu_\theta(x_t, t)$, they propose to predict the noise $\epsilon$, expressing $\mu_\theta(x_t, t)$ as
\begin{equation}
\label{eq:meanReparametrized}
    \mu_\theta(x_t, t) = \frac{1}{\sqrt{\alpha_t}}(x_t-\frac{\beta_t}{\sqrt{1-\Bar{\alpha}_t}}\epsilon_\theta(x_t, t)).
\end{equation}
This formulation allows for a simpler objective function, which is a reweighted form of the variational lower bound:
\begin{equation}
\label{eq:lossSimple}
L_{simple} = E_{t, x_0, \epsilon}[||\epsilon-\epsilon_\theta(x_t,t)||^2].
\end{equation}
As this formulation gives no learning signal for the variances $\Sigma_\theta(x_t,t)$, Ho \etal~\cite{ddpm} propose to fix $\Sigma_\theta(x_t,t)$ to time-dependent constants ${\sigma_t^2=\beta_t}$.
We can then train the model by randomly sampling a timestep $t$, applying \cref{eq:forwardSample} to get a noised sample $x_t$, use $(x_t, t)$ as input to our model to predict $\epsilon_\theta(x_t, t)$ and optimize the loss in \cref{eq:lossSimple}.

To sample from the trained model, we first generate a sample ${x_T\sim\mathcal{N}(0,1)}$ and we use the trained network to predict $\epsilon_\theta(x_t,t)$, thus obtaining $\mu_\theta(x_t,t)$ from \cref{eq:meanReparametrized}. We can now sample ${x_{t-1}\sim\mathcal{N}(x_{t-1};\mu_\theta(x_t,t);\sigma_t^2\mathbf{I})}$. This process is repeated recursively, eventually obtaining $x_0$, the generated output of our pipeline.

\section{Guidance}
\label{sec:guidance}
We apply classifier-free guidance~\cite{cfreeGuidance} to guide our model. To this end, we jointly train a conditional and an unconditional model, then we make predictions by combining their score estimates to step toward the guidance direction. To jointly train a conditional and an unconditional model, we replace with probability $p$ the input conditioning with a null token $\emptyset$. At inference time, we combine the conditional and unconditional predictions as
\begin{equation}
   \medmath{\hat{\epsilon}_\theta(x_t, t|y) = \epsilon_\theta(x_t, t|\emptyset) + w\cdot\left(\epsilon_\theta(x_t, t|y)-\epsilon_\theta(x_t, t |\emptyset) \right)},
\end{equation}
where y is the guidance token and $w\geq1$ is the guidance scale. We apply classifier-free guidance with $p=0.1$ on image tokens and CISP embeddings independently.
\subsection{Guidance scale}
In the original paper~\cite{cfreeGuidance} and in GLIDE~\cite{GLIDE}, both dealing with text-to-image generation, it was shown that the guidance scale balances between quality and diversity: higher guidance scales produce samples of higher quality and more coherent to the conditioning tokens, thus reducing the sample diversity. However, in~\cite{DALL-E2} the authors find the guidance scale to only affect photorealism (e.g., lighting, transparency) and not diversity, thus obtaining images correctly matching the input text across different guidance scales. We also find that the diversity of our generated shapes is independent of the guidance scale, although in our case 1\nobreakdash-NNA and visual realism also remain similar. This could be due to the nature of our 3D representation. Indeed, voxels only represent the shape of an object and do not consider complex aspects such as textures and light effects. Thus, if low guidance scales already obtain coherent and realistic shapes, we do not expect that more relevant details can be added when increasing the guidance scale.

\begin{table}[t]
    \centering
    \begin{tabular}{@{}l l c c@{}}
    \toprule
        ~ & ~ & \multicolumn{2}{r}{1-NNA(\%)}\\
        Shape & Model & CD & EMD \\ 
        \midrule
        \multirow{6}{*}{Aeroplane} & Unguided & 62.78 & 61.13 \\ 
        & Class-guided & 63.23 & 61.95 \\ 
        & CISP & 75.12 & 66.94 \\ 
        & Null token + null token & 58.93 & 56.93 \\
        & CISP + null token & 58.11 & 54.47 \\ 
        & CISP + $E_c$ (IC3D) & \textbf{57.64} & \textbf{53.89} \\
        \midrule
        \multirow{6}{*}{Car} & Unguided & 58.17 & 57.01 \\ 
        & Class-guided & 57.25 & 56.37 \\ 
        & CISP & 62.23 & 58.69 \\ 
        & Null token + null token & 53.20 & 53.11 \\
        & CISP + null token & 53.35 & 52.04 \\ 
        & CISP + $E_c$ (IC3D) & \textbf{52.44} & \textbf{51.68} \\ 
        \midrule
        \multirow{6}{*}{Chair} & Unguided & 70.34 & 71.39 \\ 
        & Class-guided & 68.98 & 68.98 \\ 
        & CISP & 68.77 & 61.49 \\ 
        & Null token + null token & 53.30 & 51.97 \\
        & CISP + null token & 54.18 & 52.20 \\ 
        & CISP + $E_c$ (IC3D) & \textbf{53.58} & \textbf{51.73} \\ 
        \midrule
        \multirow{6}{*}{Average} & Unguided & 63.76 & 63.18 \\ 
        & Class-guided & 63.15 & 62.43 \\ 
        & CISP & 68.71 & 62.37 \\ 
        & Null token + null token & 55.14 & 54.00 \\
        & CISP + null token & 55.21 & 52.90 \\ 
        & CISP + $E_c$ (IC3D) & \textbf{54.55} & \textbf{52.43} \\ 
        \bottomrule
    \end{tabular}
    \caption{Extended ablation study on CISP and $E_c$. Best results are highlighted in bold.}
    \label{table:ablationtokens}
\end{table}

\subsection{Guidance tokens ablation}
Our approach leverages both CISP and $E_c$ tokens to produce shapes that are realistic and coherent to the input image. As discussed in the main paper (Sec.~5.6), training a model with only $E_c$ tokens is not effective in guiding the generation toward the input image. Differently, even though the introduction of CISP tokens enables an effective guidance, the joint use of $E_c$ and CISP tokens provides even better results, allowing superior modeling of low-level details and symmetries. 

To further investigate the role of our guidance elements, we extend here our ablation by testing our model in 3 additional settings. First, we analyze the impact of training our DDPM in an unguided and class-guided setting, as to consider these two setups as baselines for our image-guided method. In the unguided setting, we train multiple models to generate shapes without any guidance input. The class-guided scenario, instead, focuses on a single model, which is guided by a class token to generate shapes of different categories. The results, reported in \cref{table:ablationtokens}, show that unguided and class-guided models obtain scores comparable to those of CISP-only guidance. CISP-only guidance, however, allows us to drive the generation through an input image instead of a class token, resulting in a more fine control of the generated shapes. As reported in the main paper, the addition of $E_c$ at training time allows us to obtain much higher generation quality. Instead, at generation time, replacing its embeddings with the learned null token at generation time does not significantly impact the results.

\section{Additional generation metrics}
Following PVD \cite{PVD} and PointFlow \cite{PointFlow}, we also measure the results of our model on Minimum Matching Distance (MMD) and Coverage (COV). We define $S_g$ as the set of generated samples and $S_r$ as the set of reference samples (taken from the test set), with $|S_r|=|S_g|$. Minimum matching distance measures the distance from each sample in $S_r$ to its nearest neighbor in $S_g$. It is defined as
\begin{equation}
        \operatorname{MMD}(S_g,S_r) =  \frac{1}{|S_r|}\sum_{r \in S_r} \min_{g \in S_g} d(g,r),
\end{equation}
where $d(.)$ is a distance metric. As done in the literature, we consider both Chamfer Distance and Earth Mover's Distance. While MMD measures the quality of the generated samples, it does not take into account their diversity. A measure of their diversity is given by COV, widely used to complement MMD. 

COV measures the percentage of samples in $S_r$ that are matched with at least one of the samples in $S_g$. A match is defined as the nearest neighbor according to a distance function $d$. COV can then be defined as
\begin{equation}
        \operatorname{COV}(S_g,S_r) =  \frac{{|\arg\min_{r \in S_r} d(g,r)\, |\, g \in S_g}|}{|S_r|}.
\end{equation}
\begin{table}[t]
\setlength{\tabcolsep}{4pt}
    \centering
    \begin{tabular}{@{}l l c c c c@{}}
    \toprule
        & & \multicolumn{2}{c}{MMD$\downarrow$} & \multicolumn{2}{c}{COV(\%)$\uparrow$}\vspace{.2em}\\
        Shape & Model & CD & EMD & CD & EMD \\ 
        \midrule
        \multirow{7}{*}{Airplane} & Shape-GF\cite{ShapeGF} & 2.703 & 0.659 & 40.74 & 40.49 \\
        & DPF-Net\cite{dpfnet} & 0.264 & 0.409 & 46.17 & 48.89 \\
        & PointFlow\cite{PointFlow} & 0.224 & 0.390 & 47.90 & 46.41 \\
        & \cite{luo2021diffusion} & 0.221 & 0.385 & 49.02 & 44.98 \\
        & PVD\cite{PVD} & 0.228 & 0.380 & 48.88 & 52.09 \\
        & LION\cite{lion} & \textbf{0.219} & \textbf{0.372} & 47.16 & 49.63\\ 
        & \textbf{Ours} & 0.224 & 0.386 & \textbf{54.03} & \textbf{54.13} \\
        \midrule
        \multirow{7}{*}{Car} & Shape-GF\cite{ShapeGF} & 9.232 & 0.756 & 49.43 & 50.28 \\
        & DPF-Net\cite{dpfnet} & 1.129 & 0.853 & 45.74 & 49.43 \\
        & PointFlow\cite{PointFlow} & \textbf{0.901} & 0.807 & 46.88 & 50.00 \\
        & \cite{luo2021diffusion} & - & - & - & - \\
        & PVD\cite{PVD} & 1.077 & 0.794 & 41.19 & 50.56 \\
        & LION\cite{lion} & 0.913 & \textbf{0.752} & 50.00 & \textbf{56.53} \\ 
        & \textbf{Ours} & 1.083 & 0.812 & \textbf{50.90} & 54.93 \\ 
        \midrule
        \multirow{7}{*}{Chair} & Shape-GF\cite{ShapeGF} & 2.889 & 1.702 & 46.67 & 48.03 \\
        & DPF-Net\cite{dpfnet} & 2.536 & 1.632 & 44.71 & 48.79 \\
        & PointFlow\cite{PointFlow} & \textbf{2.409} & 1.595 & 42.90 & 50.00 \\
        & \cite{luo2021diffusion} & 2.415 & 1.564 & 49.66 & 50.22 \\
        & PVD\cite{PVD} & 2.622 & 1.556 & 49.84 & 50.60 \\
        & LION\cite{lion} & 2.640 & 1.550 & 48.94 & \textbf{52.11} \\ 
        & \textbf{Ours} & 2.492 & \textbf{1.573} & \textbf{51.08} & 51.44 \\ 
        \bottomrule
    \end{tabular}
    \caption{Comparison of our model on generation metric against other works. The car category was not part of the evaluation of \cite{luo2021diffusion}. Best results are highlighted in bold.}
    \label{table:quantitativeResultsGeneration}
\end{table}
\cref{table:quantitativeResultsGeneration} shows MMD and COV results of our model compared to previous works. Our model is comparable to the others in MMD, which slightly suffers from the conversion from voxel to point cloud; on the other hand, it obtains significantly higher results in COV in nearly all cases. In these two cases LION~\cite{lion} performs slightly better, however our model is the second best and still competitive to it.

\begin{table}[t]
    \centering
    \begin{tabular}{@{}l c c c c c c@{}}
        \toprule
        & \multicolumn{2}{r}{1-NNA(\%)} & \multicolumn{2}{c}{MMD$\downarrow$} & \multicolumn{2}{c}{COV(\%)$\uparrow$}\\
        Shape & CD & EMD & CD & EMD & CD & EMD \\ 
        \midrule
        Table      & 55.63 & 53.89 & 2.326 & 1.573 & 53.85 & 54.10\\
        Watercraft & 56.95 & 54.03 & 1.984 & 1.453 & 53.57 & 52.50\\
        \bottomrule
    \end{tabular}
    \caption{Generation performance of IC3D on additional object categories.\vspace{-1em}}
    \label{table:additionalCategories}
\end{table} 
\begin{figure}[t]
  \centering
   \includegraphics[width=0.9\linewidth]{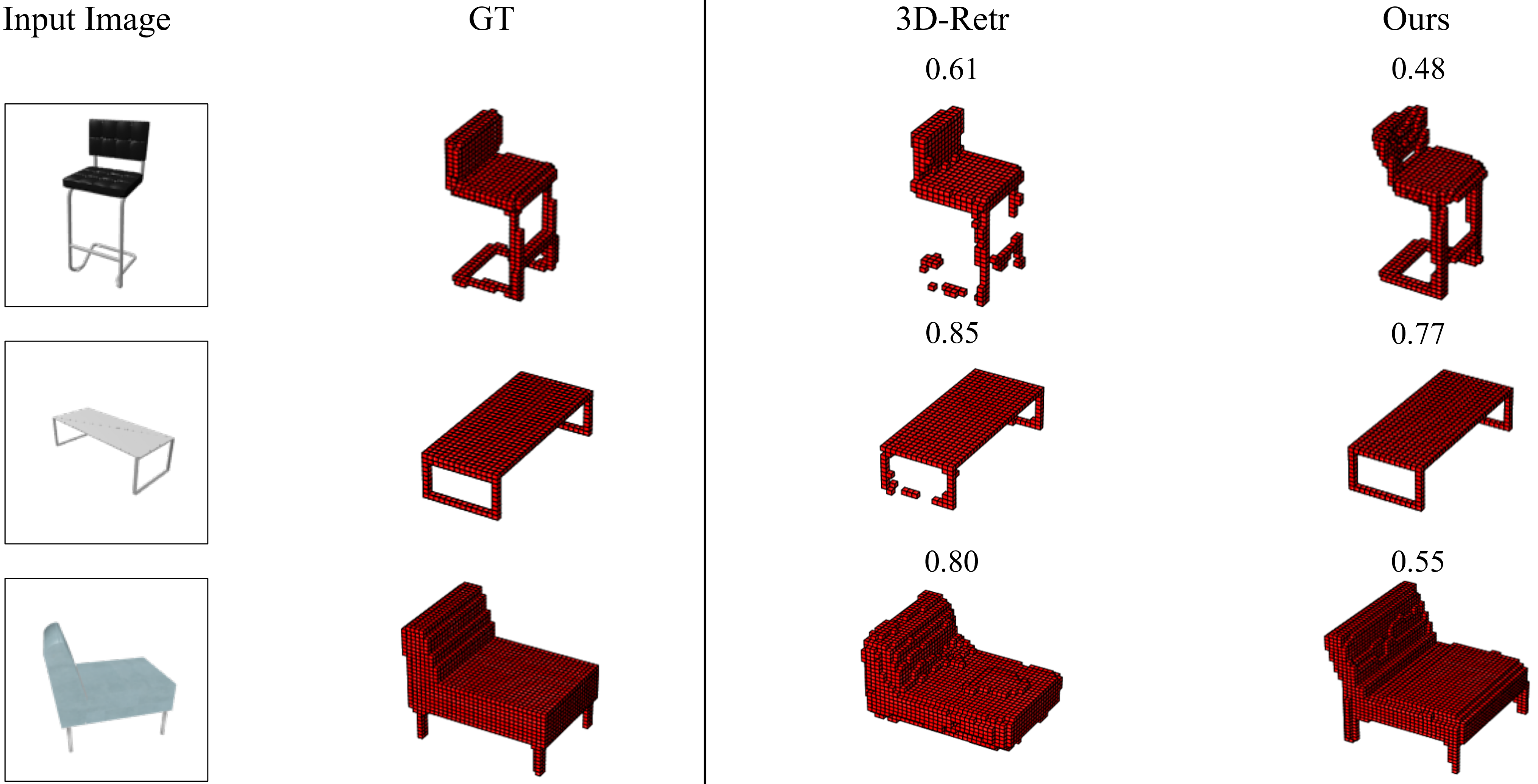}
   \caption{Even if our model produces coherent and realistic shapes, IoU favors the shapes reconstructed by 3D-RETR~\cite{3d-retr}, which may contain structural integrity problems. Although this image is already reported in the main paper (Fig.~10), we report it here for ease of discussion.\vspace{-1em}}
   \label{fig:flawsIoU}
\end{figure}
\begin{figure}[t]
  \centering
   \includegraphics[width=0.9\linewidth]{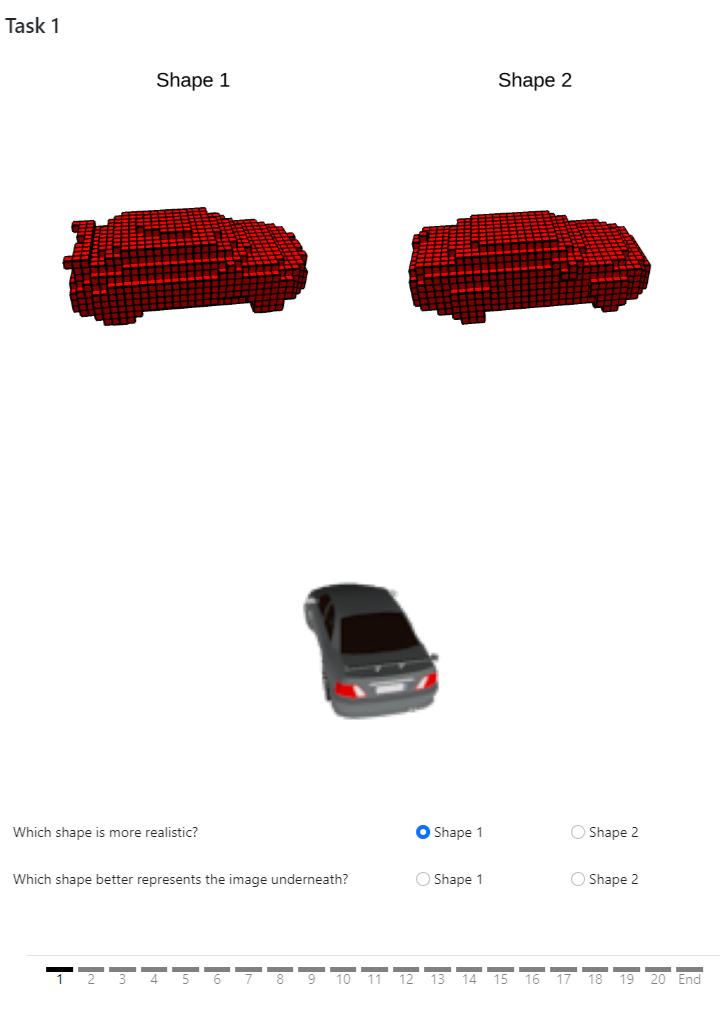}
   \caption{Human evaluation interface. The query image is shown only after the first question has been answered. The second question is unlocked togheter with the image.\vspace{-1em}}
   \label{fig:heInterface}
\end{figure}
\section{Generation metrics for additional categories}
We report in \cref{table:additionalCategories} our results on the table and watecraft category, on which generative models in the literature do not compare.
\section{Considerations against standard reconstruction metrics}
As discussed in the main paper, metrics such as IoU and F-score fail to capture properties such as realism, integrity and structural correctness of 3D shapes. Indeed, they only measure the reconstruction quality by considering the number of correctly predicted voxels values. As a consequence, models trained to optimize these quantities can generate structurally wrong shapes, such as tables and chairs with discontinuous or missing legs, or cars with flat wheels. Indeed, such errors are caused by only a few wrongly predicted voxels, thus having a small impact on metrics such as IoU and F-score. Moreover, IoU is sensitive to the slightest shift or scale of the predicted shape with respect to the ground truth. This is problematic for our model, since, being generative, it may produce a conceptually correct shape but place it slightly scaled or shifted compared to the ground truth. Examples of these scenarios are highlighted in \cref{fig:flawsIoU}. Although our model produces realistic and structurally intact shapes, they may exhibit different details and be slightly scaled or shifted. This is due to the probabilistic approach and heavily impacts our overall score. These considerations led us to perform a human evaluation, reported in the main paper, in order to evaluate both our coherence to the query image and the realism of our shapes from a human perspective.

\begin{table*}[th!]
    \centering
    \begin{tabular}{@{}l c c c c c c c@{}}
    \toprule
        ~ & \multicolumn{7}{c}{Votes for IC3D (\%)}\\
        Category & 0/5 & 1/5 & 2/5 & 3/5 & 4/5 & 5/5 & 3/5 or higher\\
        \midrule
        Aeroplane & ~6.00 & 16.50 & 16.00 & 24.50 & 19.50 & 17.50 & \textbf{61.50} \\
        Car & 12.50 & 9.50 & 19.00 & 23.50 & 19.50 & 16.00 & \textbf{59.00} \\
        Chair & ~7.50 & 11.00 & 13.00 & 20.00 & 16.50 & 32.00 & \textbf{68.50} \\
        \midrule
        Overall & ~8.67 & 12.33 & 16.00 & 22.67 & 18.50 & 21.83 & \textbf{63.00} \\ 
        \bottomrule
    \end{tabular}
    \caption{\textbf{Coherence} per-class human evaluation results. Our model obtains the majority vote (in bold) on all object categories.}
    \label{table:perclassCoherence}
\end{table*}
\begin{table*}[th!]
    \centering
    \begin{tabular}{@{}l c c c c c c c@{}}
    \toprule
        ~ & \multicolumn{7}{c}{Votes for IC3D (\%)}\\
        Category & 0/5 & 1/5 & 2/5 & 3/5 & 4/5 & 5/5 & 3/5 or higher\\
        \midrule
        Aeroplane & ~3.50 & 12.50 & 19.00 & 16.50 & 21.00 & 27.50 & \textbf{65.00} \\
        Car & 9.50\% & 18.50 & 20.50 & 18.50 & 19.00 & 14.00 & \textbf{52.00} \\ 
        Chair & 4.00 & 4.00 & ~1.50 & ~9.50 & 20.00 & 61.00 & \textbf{91.00} \\
        \midrule
        Overall & 5.67 & 11.67 & 13.67 & 14.83 & 20.00 & 34.17 & \textbf{69.00} \\
        \bottomrule
    \end{tabular}
    \caption{\textbf{Realism} per-class human evaluation results. Our model obtains the majority vote (in bold) on all object categories.}
    \label{table:perclassRealism}
\end{table*}
\begin{table*}[th!]
    \begin{adjustbox}{width=\linewidth,center}
    \begin{tabular}{@{}lccccccccccc@{}}
    \toprule
    ~ & \thead{3D\nobreakdash-R2N2\\\cite{3dr2n2}} & \thead{OGN\\\cite{ogn}} & \thead{Pixel2Mesh\\\cite{pixel2mesh}} & \thead{IM\nobreakdash-Net\\\cite{imnet}} & \thead{{Pix2Vox++/F}\\\cite{pix2vox++}} & \thead{3D\nobreakdash-RETR\\\cite{3d-retr}} & \thead{TMVNet \\\cite{tmvnet}} & \thead{Ours\\(1)} & \thead{Ours\\(5)} & \thead{Ours\\(10)} & \thead{Ours\\(15)} \\
    \midrule
        Aeroplane & 0.512 & 0.587 & 0.508 & 0.702 & 0.607 & 0.704 & 0.691 & 0.540 & 0.600 & 0.620 & 0.630 \\
        Car  & 0.798 & 0.828 & 0.670 & 0.756 & 0.841 & 0.861 & 0.870 & 0.790 & 0.824 & 0.833 & 0.838 \\
        Chair & 0.466 & 0.483 & 0.484 & 0.644 & 0.548 & 0.592 & 0.721 & 0.407 & 0.476 & 0.494 & 0.506 \\
        \bottomrule
    \end{tabular}
    \end{adjustbox}
    \caption{Per-class IoU single-view 3D reconstruction results.}
    \label{table:perClassResultsIoU}
\end{table*}
\begin{table*}[th!]
    \begin{adjustbox}{width=\linewidth,center}
    \begin{tabular}{@{}lccccccccccc@{}}
    \toprule
    ~ & \thead{3D\nobreakdash-R2N2\\\cite{3dr2n2}} & \thead{OGN\\\cite{ogn}} & \thead{Pixel2Mesh\\\cite{pixel2mesh}} & \thead{IM\nobreakdash-Net\\\cite{imnet}} & \thead{{Pix2Vox++/F}\\\cite{pix2vox++}} & \thead{3D\nobreakdash-RETR\\\cite{3d-retr}} & \thead{TMVNet \\\cite{tmvnet}} & \thead{Ours\\(1)} & \thead{Ours\\(5)} & \thead{Ours\\(10)} & \thead{Ours\\(15)} \\
    \midrule
        Aeroplane & 0.412 & 0.487 & 0.376 & 0.589 & 0.583 & 0.612 & 0.594 & 0.492 & 0.543 & 0.561 & 0.570 \\
        Car  & 0.481 & 0.514 & 0.486 & 0.304 & 0.564 & 0.511 & 0.602 & 0.394 & 0.425 & 0.433 & 0.439 \\ 
        Chair & 0.238 & 0.226 & 0.386 & 0.442 & 0.309 & 0.352 & 0.520 & 0.203 & 0.238 & 0.248 & 0.255 \\ 
        \bottomrule
    \end{tabular}
    \end{adjustbox}
    \caption{Per-class F-score single-view 3D reconstruction results.\vspace{-1em}}
    \label{table:perClassResultsfscore}
\end{table*}

\begin{figure*}[t]
  \centering
   \includegraphics[width=0.65\linewidth]{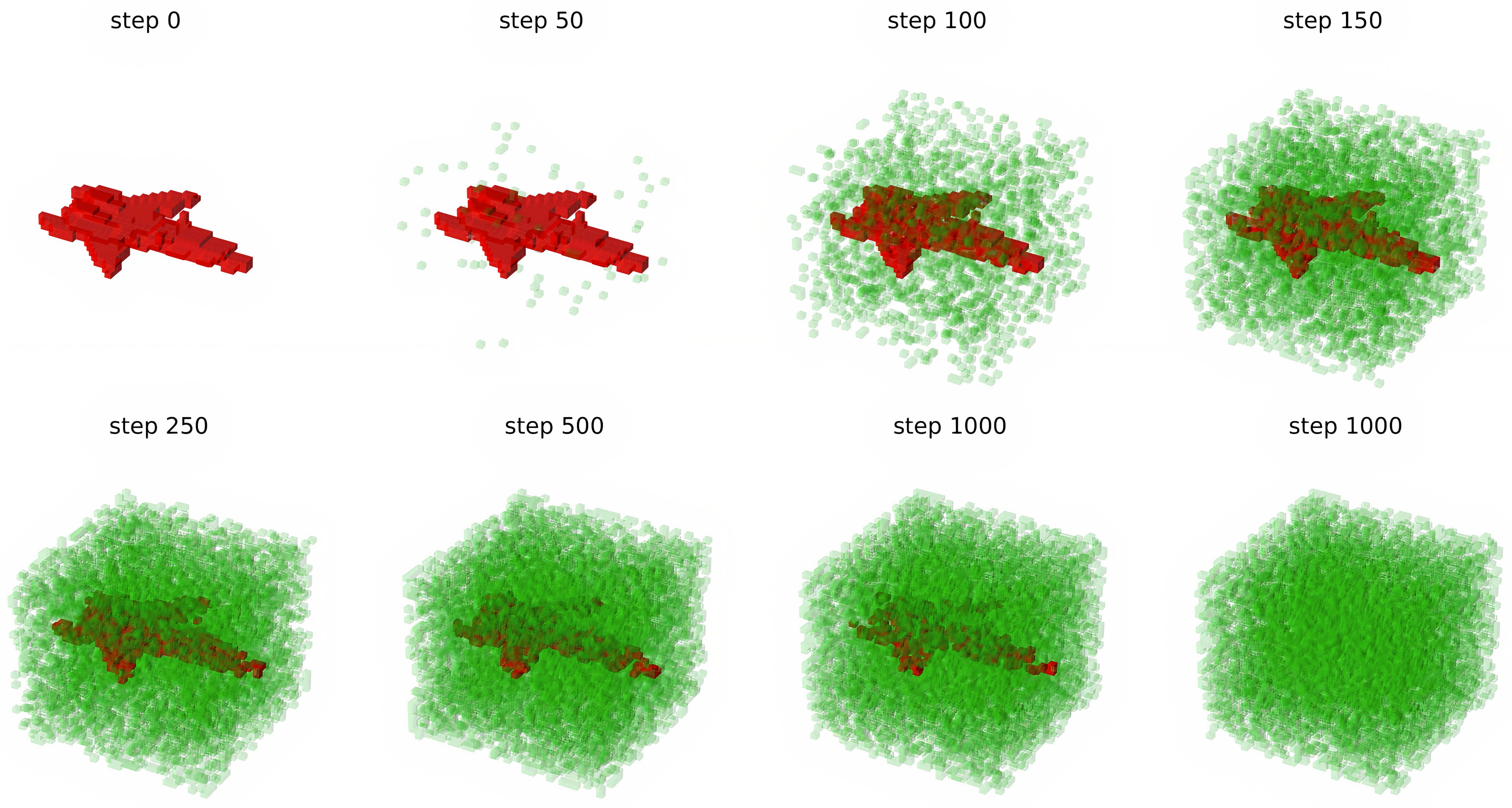}
   \caption{Voxel diffusion process. At each diffusion step, noise is added to the volumetric data, and samples are thresholded at 0.5 to visualize binary data. Originally filled values are shown in red for comparison. Step 1000 is displayed also without red highlighting, showing how the original shape is completely lost. }
   \label{fig:diffVoxelProcess}
\end{figure*}
\begin{figure*}[th!]
  \centering
   \includegraphics[width=0.85\linewidth]{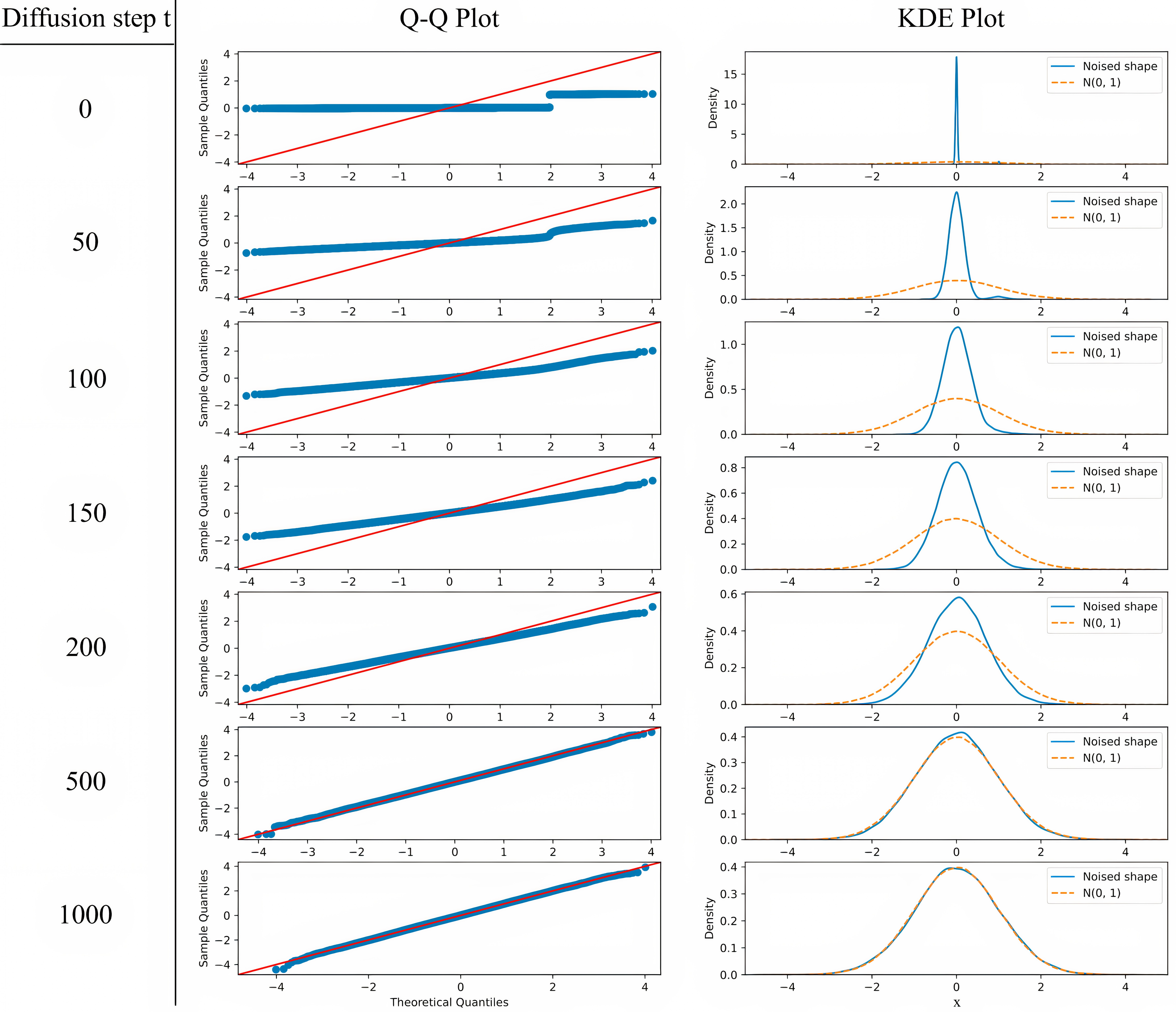}
   \caption{Distribution of values throughout the forward process. We observe that the diffusion process progressively transforms the initial distribution into a standard Normal distribution.}
   \label{fig:diffVoxelNormalAnalysis}
\end{figure*}

\section{Human evaluation additional material}
\noindent\textbf{Interface details}
We report additional details on the human evaluation procedure.
\cref{fig:heInterface} shows the interface presented to our human evaluators. To ensure that the evaluators would not be biased by our interface, we adopted the following precautions: (1) the shapes generated by our model and by 3D-RETR are placed to the left or to the right randomly. (2) the GIFs are synchronized to allow for a direct comparison of details. (3) the query image is centered to avoid a closeness bias. 
Throughout our evaluation campaign, we recorded only the anonymized answers of our evaluators. We did not record any personal information, and we made this clear to each participant.

\noindent\textbf{Per-class results}
We report the per-class human evaluation results in \cref{table:perclassCoherence,table:perclassRealism}. Human evaluators prefer shapes generated by our model for both coherence and realism in all categories. Notice how the scores obtained by our model for the chair class are the highest among all categories. Our model achieves the majority of the votes for realism in 91\% of the tested shapes. For cars, instead, we register the lower margin, while still being preferred by the evaluators. We think this is because cars are characterized by less complex shapes, thus both models produce very similar results, forcing the evaluators to chose almost randomly between the two. Single-view 3D reconstruction results (\cref{table:perClassResultsIoU,table:perClassResultsfscore}) also support this hypothesis, as, for most models, cars are the best reconstructed class, while chairs are the worst. In fact, among the three tested categories, chairs display the most complex structure, as they often present thin elements. For example, legs and  armrests are often modeled incompletely or incorrectly by reconstruction pipelines. Our model IC3D, instead, thanks to its generative nature, is able to correctly produce structurally realistic samples. 

\section{Analyzing voxel diffusion}
As voxel DDPMs have never been deployed before, we study the effect of the diffusion process with Gaussian transitions on voxels. In fact, being voxels binary data, we may ask if a Gaussian process is effectively able to gradually destroy the information in a sample, which is crucial for the reverse step. 

We study the distribution of volume values across different steps in the forward diffusion process, and show the results in \cref{fig:diffVoxelProcess,fig:diffVoxelNormalAnalysis}. 

Specifically, \cref{fig:diffVoxelProcess} visually shows the effect of the forward process on a binarized shape. We highlight in red voxels with value 1 belonging to the original shape and in green voxels with value 1 which are not from the original shape. We can observe both how originally filled voxels are erased by the process and how new voxels are filled. The result of the last diffusion step is displayed with and without highlighting, to show that information about the original shape is lost. 

In \cref{fig:diffVoxelNormalAnalysis}, we analyze the distribution of the data, comparing it to a standard Normal distribution at different timesteps by computing Quantile-Quantile (QQ) and Kernel Density Estimation (KDE) plots. 
QQ plots assess the plausibility that a set of data is coming from a theoretical distribution (in our case, a standard Normal distribution) by plotting their sets of quantiles against one another. Data coming from the theoretical distribution give origin to a straight line, as in our case. Instead, KDE plots allow us to visualize the distribution of our data by estimating its density function. \cref{fig:diffVoxelNormalAnalysis} shows QQ and KDE at different timesteps, proving how the sample distribution progressively approximates a standard Normal distribution.

{\small
\bibliographystyle{IEEEtran}
\bibliography{bib}
}